\documentclass[letterpaper, 10 pt, conference]{ieeeconf}  
\IEEEoverridecommandlockouts

\usepackage{cite}
\usepackage{amsmath,amssymb,amsfonts}
\usepackage{algorithmic}
\usepackage{graphicx}
\usepackage{textcomp}
\usepackage{xcolor}
\def\BibTeX{{\rm B\kern-.05em{\sc i\kern-.025em b}\kern-.08em
    T\kern-.1667em\lower.7ex\hbox{E}\kern-.125emX}}

\usepackage{float}
\usepackage{tikz}

\usepackage[breaklinks=true,bookmarks=false]{hyperref}

\newcommand{\cutpic}[5]{
    \begin{tikzpicture} 
       \begin{scope}
            \clip [rounded corners=.40cm] (0,0) rectangle coordinate (centerpoint) ++(2.9cm,2.9cm); 
            \node [inner sep=0pt] at (centerpoint) {\includegraphics[trim=#1 #2 #3 #4,clip,width=0.2\linewidth]{#5}};
        \end{scope}
    \end{tikzpicture}
}

\begin{document}



\author{Frederik Hagelskjær and Rasmus Laurvig Haugaard
\thanks{
Both authors are with SDU Robotics, Mærsk Mc-Kinney Møller Institute, University of Southern Denmark, 5230 Odense M, Denmark
        {\tt \{frhag,rlha\}@mmmi.sdu.dk}}%
}

\title{KeyMatchNet: Zero-Shot Pose Estimation in 3D Point Clouds by Generalized Keypoint Matching}

\maketitle


\begin{abstract}
In this paper, we present KeyMatchNet, a novel network for zero-shot pose estimation in 3D point clouds. Our method uses only depth information, making it more applicable for many industrial use cases, as color information is seldom available. The network is composed of two parallel components for computing object and scene features. The features are then combined to create matches used for pose estimation. The parallel structure allows for pre-processing of the individual parts, which decreases the run-time. 

Using a zero-shot network allows for a very short set-up time, as it is not necessary to train models for new objects.
However, as the network is not trained for the specific object, zero-shot pose estimation methods generally have lower accuracy compared with conventional methods. To address this, we reduce the complexity of the task by including the scenario information during training. This is typically not feasible as collecting real data for new tasks drastically increases the cost. However, for zero-shot pose estimation, training for new objects is not necessary and the expensive data collection can thus be performed only once. 

Our method is trained on 1,500 objects and is only tested on unseen objects. We demonstrate that the trained network can not only accurately estimate poses for novel objects, but also demonstrate the ability of the network on objects outside of the trained class. Test results are also shown on real data. We believe that the presented method is valuable for many real-world scenarios. 
Project page available at \href{https://keymatchnet.github.io}{keymatchnet.github.io}
\end{abstract}


\begin{figure}[t]
\begin{center}
   \includegraphics[width=0.75\linewidth]{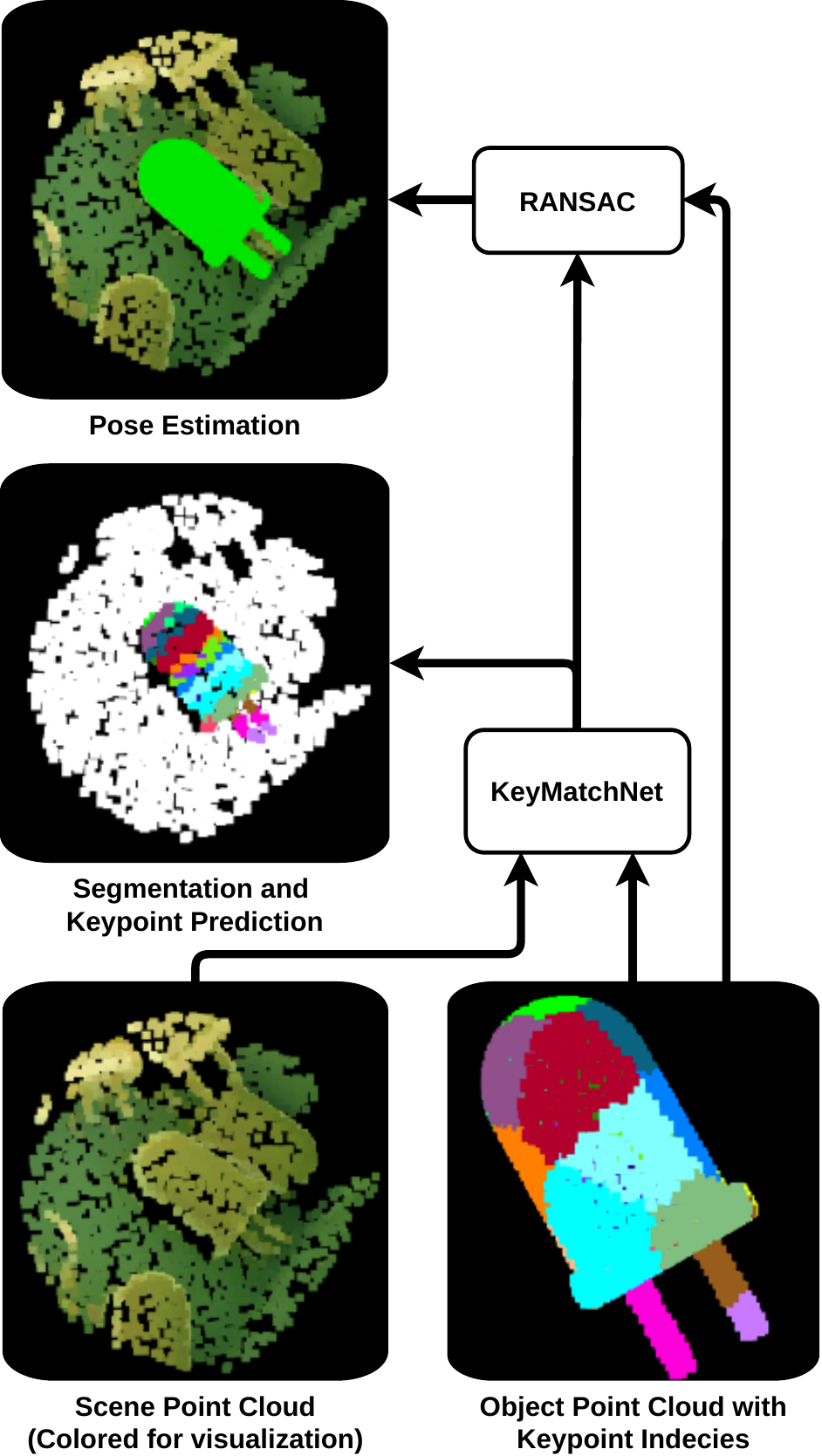}
\end{center}
   \caption{Illustration of the pose estimation method on an unseen object. 
The input to the network is a scene point cloud and an object point cloud. The object keypoints are visualized by the different color segments. The network output is both instance segmentation and keypoint predictions, which are combined to provide predictions only for the object. Finally these predictions are used in RANSAC \cite{fischler1981random} for pose estimation.
The striped keypoint prediction pattern is the result of the objects' rotational symmetry. 
   }
  \vspace{-6mm}
\label{fig:example}
\end{figure}

\section{Introduction}
Pose estimation enables greater flexibility in robotics as new objects can be manipulated without the need for mechanical fixtures or teaching specific robot positions. This enables shorter changeover times and faster adaptation to production demands. However, the set-up of computer vision algorithms can itself be a very time-consuming task \cite{hagelskjaer2018does}. There is, therefore, great interest in pose estimation solutions with simple set-ups. Deep learning has allowed learning the specifics of the object and the scenario, thus giving much better performance than human fine-tuning \cite{hodavn2020bop, sundermeyer2023bop}. However, collecting the data for training the deep neural networks can be very time-consuming, thus limiting the usability. To avoid this data collection task, synthetic data has gained widespread use \cite{hodavn2020bop}. But generating large amounts of data, and then training the network for each new object is computationally expensive and increases set-up time. To address these problems, we introduce KeyMatchNet, a neural network for zero-shot pose estimation.

A feature of KeyMatchNet is that it only uses depth information and exclude color. To the best of our knowledge this is the first colorless zero-shot network for pose estimation. 
The color information is omitted as most industrial objects does not include detailed surface information. Thus by omitting color our method is also usable for real industrial use cases. This is in contrast to most other methods that, does obtain high scores on benchmark datasets, but are not usable for many real objects. 

%

Another important aspect of our developed network is reusability. 
The reusability consists of two parts. Firstly, rather than training a single model for each object, a zero-shot pose estimation algorithm is built with a single network for all objects. This is accomplished by learning to match keypoints from the object to the scene, instead of learning specific features from the object. Thus, the network input is both scene and object information. For a novel object, the network can be reused without re-training. Additionally, compared with similar methods \cite{gou2022unseen} our network is split into parallel computation of object- and scene-features. This allows for pre-computing the object features, giving a significant speed-up at run-time. Additionally, if different objects are pose estimated the scene feature can be reused.
%

Zero-shot pose estimation methods are, generally, less accurate compared with methods trained for specific objects \cite{labbe2022megapose}, as these methods can integrate object information into the model. However, these methods often use synthetic training data and, therefore, do not integrate real scene information. 
As zero-shot methods do not require training for new objects, it is much more feasible to use real training data and integrate scene information into the network. 

For many robotic set-ups this much better fits the application, i.e. novel objects, but unchanging scene. New objects can be introduced faster as training is not required, and the power consumption for training will be removed. 
In this paper we focus on the task of bin picking with homogeneous bins, which is a difficult challenge that often occurs in industry \cite{yokokohji2019assembly}. The homogeneous bin also removes the need for object detection and allows us to only focus on pose estimation. We recognize that the current approach as shown in e.g., \cite{sundermeyer2023bop} has huge importance, but also state it is not the best solution for all tasks. In this paper we show that generalized pose estimation can obtain very good performance when restricting the scenario. We believe that these results invite further research into this topic, as creating flexible set-ups with lower energy consumption is an important topic. 
In this work the following contributions are presented.
\begin{itemize}
    \item A zero-shot pose estimation method for colorless point clouds.
    \item A parallel structure allowing pre-processing of individual parts.
    \item A dataset of 1,500 objects in homogeneous bins.
    \item Demonstrating the feasibility of zero-shot pose estimation for industrial use cases.
\end{itemize}



\section{Related Works}
The importance of visual pose estimation along with the complexity has resulted in a large amount of different solutions \cite{hodan2018bop,hodavn2020bop,sundermeyer2023bop}. In this section an overview will be given of classic pose estimation methods, deep learning based methods and current zero-shot pose estimation methods.

\subsection{Classic Pose Estimation} 
In the classic pose estimation case two different approaches are generally used, template matching and feature matching. Template matching \cite{ulrich2012combining, hinterstoisser2012model} is performed using a cascade search of where templates of the object are matched at different positions in the image. To perform the search, images have to be generated of all poses that the object can appear in.

In feature matching pose estimation is based on matching features between the scene and object, and computing the pose by e.g. Kabsch-RANSAC \cite{kabsch1976solution,fischler1981random} or pose voting \cite{buch2017rotational}. The matches are computed using handcrafted features, which are generally computed in 3D point clouds. A huge amount of handcrafted features have been developed \cite{guo2016comprehensive}, with Fast Point Feature Histograms (FPFH) \cite{rusu2009fast} being one of the best performing features.
The benefit of the classic method is that the set-up can be performed using only the CAD model. However, these methods are weak towards clutter and occlusion, and thus are often not usable in real scenarios.

\subsection{Deep Learning Based} Generally deep learning based methods are based on color information \cite{hodavn2020bop, sundermeyer2023bop}. This is possibly a result of many deep learning based methods developed for this space, with huge pre-trained networks available. These methods have vastly outperformed the classical methods, however, a network is often trained per object \cite{hodavn2020bop, sundermeyer2023bop}. 
Deep learning for pose estimation has also been performed in point clouds with methods such as PointVoteNet \cite{hagelskjaer2020pointvotenet}. We base our method on PointVoteNet, but only train a single network for all objects.

\noindent \textbf{2D methods:}
SSD-6D \cite{kehl2017ssd} is similar to template matching methods, in that it employs the cascade search. However, unlike template matching the comparison is performed using a neural network. The network is trained to classify the presence and correct orientation of an object. Thus by cascade search in the image the correct object pose is found. The position and orientation can also be computed independently \cite{xiang2017posecnn, li2019cdpn}. In PoseCNN \cite{xiang2017posecnn} the orientation is computed using a regression network.

An alternate approach is by detecting keypoints of the object \cite{tekin2018real, li2019cdpn}. This is performed in  BB-8 \cite{rad2017bb8} where the bounding box is predicted and then used for pose estimation. PVNet \cite{peng2019pvnet} compute keypoint locations by first segmentation the object, and then computing the relative position of keypoint for all pixels belonging to the object.

A method more similar to our approach is EPOS  \cite{hodan2020epos} where both object segmentation and dense keypoint predictions are calculated. This approach is also performed in DPOD \cite{zakharov2019dpod} which also employs pose refinement. Unlike our methods these methods use RGB information and train a single network per object.


\begin{figure*}[t]
    \vspace{1.5mm}
    \begin{center}
       \includegraphics[angle=90, width=0.97\linewidth]{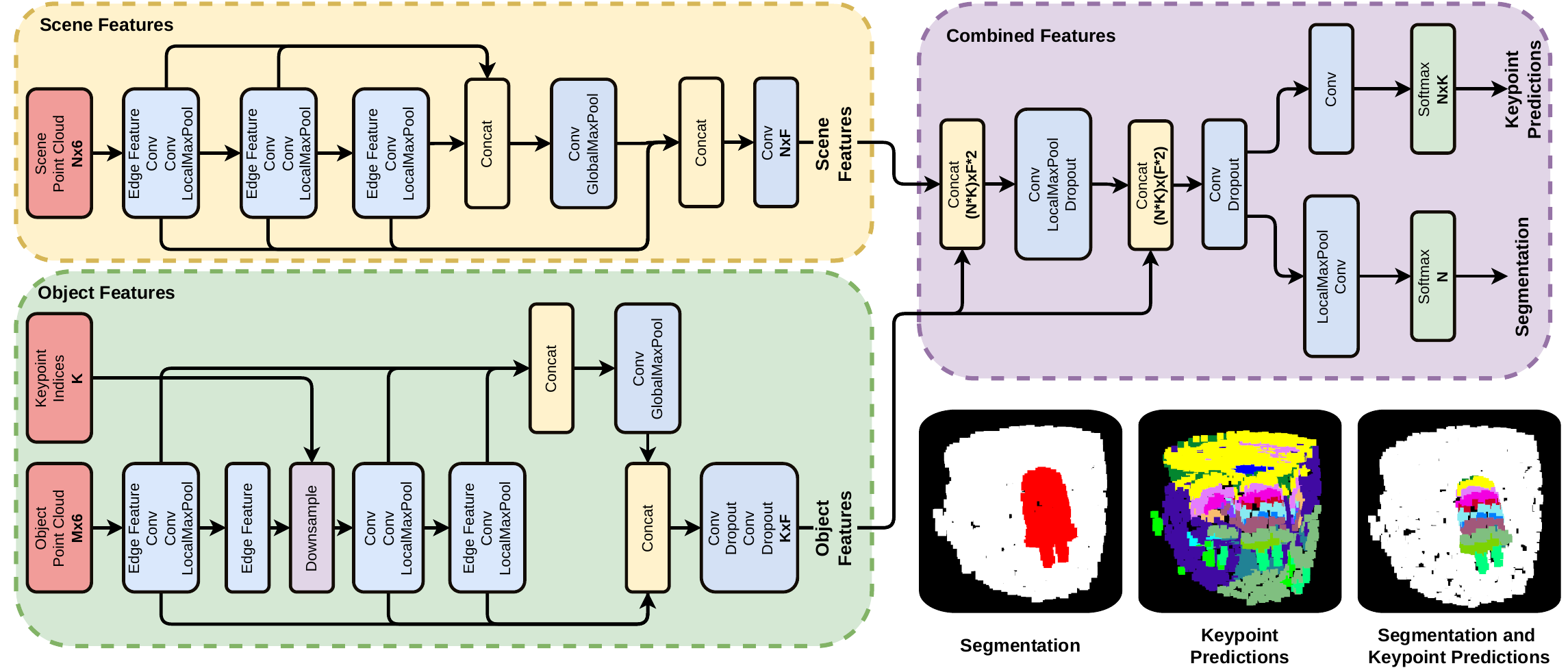}
	    \caption{The network structure of the developed method. Bold text indicates the matrix size, where \textbf{N} is the number points in the scene, \textbf{M} is the number of points in the object, \textbf{K} is the number of keypoints, and \textbf{F} is the feature size at each point. The input point clouds consist of the combined xyz and normal vector information and are thus size \textbf{6}. Object and scene features are computed independently, which allow for precomputed object features. The Segmentation and Keypoint Prediction outputs are shown along with the combination into matches. 
	    } 
        \label{fig:network}
        \vspace{-6mm}
    \end{center}
\end{figure*}

\noindent \textbf{3D methods: } DenseFusion \cite{wang2019densefusion} is a method that combines both 2D and 3D information. Initial segmentations are found in 2D, and features computed in 2D are then integrated with 3D features. The 3D features are computed using PointNet \cite{qi2017pointnet}. Finally a PnP is used to perform pose estimation using keypoints from the network.
PVN3D \cite{he2020pvn3d} is another method that combines 2D and 3D features. It also computes keypoints for which are used for pose estimation. Unlike our method the keypoints are the object bounding box, and not keypoints on the object.

A method more similar to our method is PointVoteNet \cite{hagelskjaer2020pointvotenet}. Here PointNet \cite{qi2017pointnet} is used to compute keypoints for each scene point. Similar to our method the computation is also performed without color information. In the extended version  \cite{hagelskjaer2021bridging} DGCNN \cite{dgcnn} is used and the segmentation and feature computation is separated, both of which appears in our method. ParaPose \cite{hagelskjaer2022parapose} use this method combined with automatic parameter to obtain state-of-the-art performance on the Occlusion dataset \cite{brachmann2014learning}. However, dissimilar to our method, for PointVoteNet a network is trained for each object.

\subsection{Generalized Pose Estimation} 
Several approaches have been developed for generalized pose estimation.
As in the single object based pose estimation, the field of generalized pose estimation is also dominated by color-based methods \cite{shugurov2022osop, labbe2022megapose, he2022fs6d}. The general approach for these methods is to match templates of the object with the real image, similar to SSD-6D. These templates can either be generated synthetically as in \cite{nguyen2022templates} or with a few real images as in FS6D \cite{he2022fs6d}. The same approach have also been used for tracking of unknown objects \cite{nguyen2022pizza}. 
As opposed to SSD-6D the network does not learn to recognize specific poses of objects, but instead to compare how well an image match the real scene.
By generating synthetic views of novel objects, these methods are thus able to perform pose estimation without training new networks.
MegaPose6D \cite{labbe2022megapose} is a notable example where the network is trained on a huge dataset with 2 million images.

The method most similar to ours is a point cloud based method \cite{gou2022unseen}. It also uses the object point cloud as input along with the scene point cloud. 
The method also employs a segmentation step as in our method, however the pose estimation  part of the method whereas our approach matches specific keypoint.
Several other differences are also present compared with our approach; the method includes color information which is often much more difficult to obtain. It does not limit the scenario and thus does not obtain the increased performance from this. And it does not separate the object and scene features, thus the object features must be computed for each pose estimation.

\section{Method}
%
The developed method is a network that matches scene-points to object keypoints. These matches are then fed into a pose estimation solver, in our case Kabsch-RANSAC \cite{kabsch1976solution,fischler1981random}, and a final pose is found. 
The scene and object data are point clouds without color information.
Color information is not used as it is seldom available for CAD (Computer Aided Design) models and would limit the usability of the method.
The network structure is made with two parallel components computing both scene- and object-features independently. The network structure is shown in Fig.~\ref{fig:network}. 
The scene features are computed for all points using the standard DGCNN \cite{wang2019dynamic} structure. However, for the object only the keypoint features are computed.
Thus, after the second neighbor computation in the DGCNN, the object point cloud is down-sampled to only contain the keypoints.
To ensure that the all keypoints contain knowledge about all other keypoint features, the number of neighbors is set to match the number of keypoints.


After both object- and scene-features are computed, the features are combined. Pairs of object-scene-feature vectors are created concatenating each scene-feature with each object-feature.
The resulting number of pairs is the number of scene-points and key-points multiplied.
%
%

An MLP (Multi Layer Perceptron) then processes each scene-keypoint pair independently and a local maxpool combines all keypoint information at the scene point.

This new scene feature is again combined with the keypoint features, and processed by an MLP.
Two different MLPs are then used to compute the segmentation and the keypoint predictions.
To compute the segmentation a local maxpool is applied, followed by an MLP which gives segmentation for each point. 
To compute the keypoint matches an MLP is applied to the combined features 
giving a score for each scene-keypoint pair. 
A softmax is then applied across the keypoint domain for each scene point.

Finally, Kabsch-RANSAC is used with the segmentation and keypoint predictions to compute the object pose. We employ the vote threshold as in \cite{hagelskjaer2020pointvotenet} to allow multiple keypoint matches at a single scene point. Compared with the vote threshold of 0.95 of \cite{hagelskjaer2021bridging}, our threshold is set at 0.7 as the zero-shot method is expected to be less precise.



\subsection{GPU based RANSAC:}
To compare our developed method with classic pose estimation methods the RANSAC implemented in Open3D \cite{Zhou2018} is used. ICP \cite{arun1987least} is then used for refinement. However, as the RANSAC method is running on the CPU a significant increase in run-time occurs. To decrease the run-time a RANSAC has also been implemented on the GPU using PyTorch \cite{Paszke_PyTorch_An_Imperative_2019}. 

The GPU based RANSAC is implemented as a Coarse-to-Fine RANSAC, with five iterations. Initially, 1000 poses are computed using the Kabsch \cite{kabsch1976solution} algorithm on triplets sampled from the matches. The poses are then sorted according to the amount of inlying matches. An additional requirement is that the normal vector between matches cannot be larger than 30 degrees. This ensures that only the parts of the object pointing towards the camera are matched.


The best scoring pose is then selected, and further refinement is performed. The pose is refined by computing Kabsch using only matches within the inlier distance. By gradually decreasing the inlier distance the pose is iteratively refined. 
As this method includes refinement, ICP \cite{arun1987least} is not used for the GPU based RANSAC.


\subsection{Generating object data:}
The object point clouds is generated using Poisson sampling to obtain 2048 evenly sampled points on the surface. Farthest point sampling is then used to obtain the keypoints spread evenly on the object.
%

\subsection{Computing object features offline:}
As shown in Fig.~\ref{fig:network} the features computed from the object point cloud are independent of the features computed from the scene point cloud. This is opposed to \cite{gou2022unseen}. This allows us to pre-compute the object features, reducing the run-time and the computational cost.
During training the keypoints are continuously re-sampled, to avoid the network over-fitting to a specific combination.


\subsection{Generating scene data:}
\begin{figure}[t]
    \vspace{1.5mm}
\begin{center}
   \includegraphics[width=0.95\linewidth]{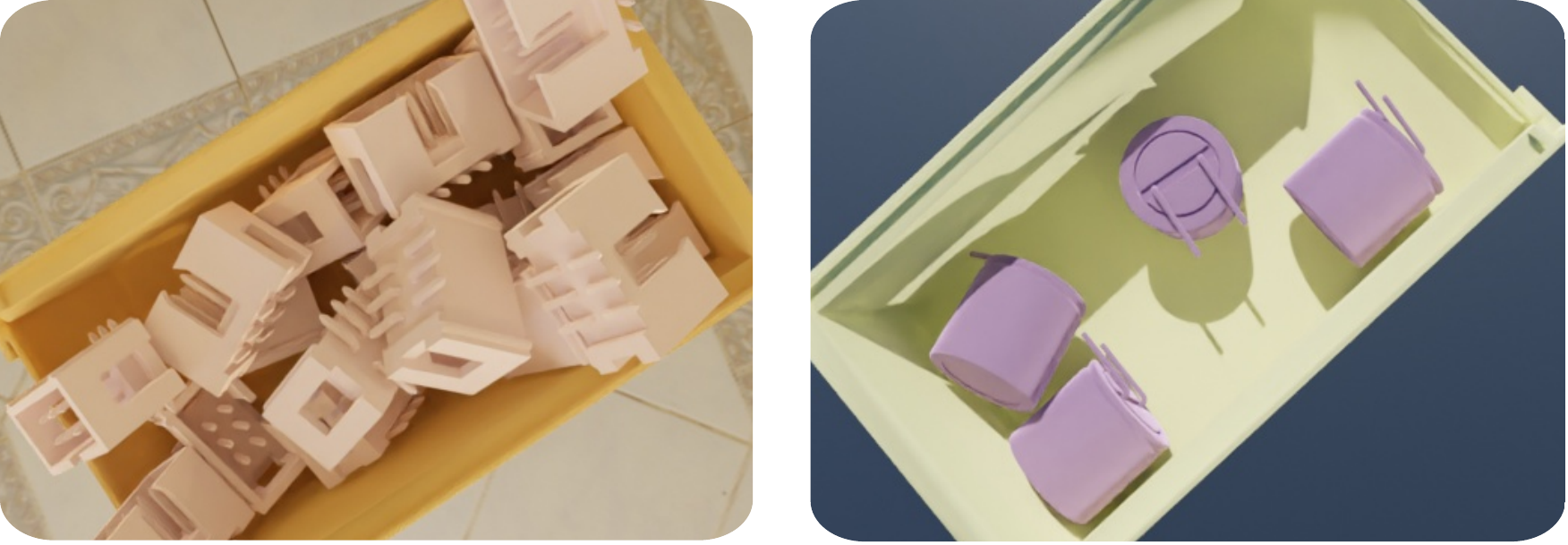}
\end{center}
   \caption{Examples of the training data. Colors are only for visualization.}
\label{fig:dtrainata}
    \vspace{-4mm}
\end{figure}

The scene point clouds are generated using BlenderBin\footnote{\url{https://github.com/hansaskov/BlenderBin/}}. BlenderBin is an open-source tool built using the BlenderProc \cite{denninger2019blenderproc} simulator. It allows to easily generate synthetic data with objects placed in bins. For each object images are created with the number of objects in the bin ranging from one to twenty. The bin model is kept the same in all scenes. Examples of the training data is shown in Fig.~\ref{fig:dtrainata}.

As the scenario is homogeneous bin-picking the detection is vastly simplified. By segmenting the known bin, all remaining points belong to objects. As the contents are homogeneous any random point is known to belong to a correct object. 
By then extracting a point cloud around the sampled point using the object diagonal, all points in the scene belonging to the object will be obtained. The point cloud is then centered around the sampled point to allow for instance segmentation. 

\begin{figure*}[htb]
    \vspace{1.5mm}
\begin{center}
   \includegraphics[width=0.9\linewidth]{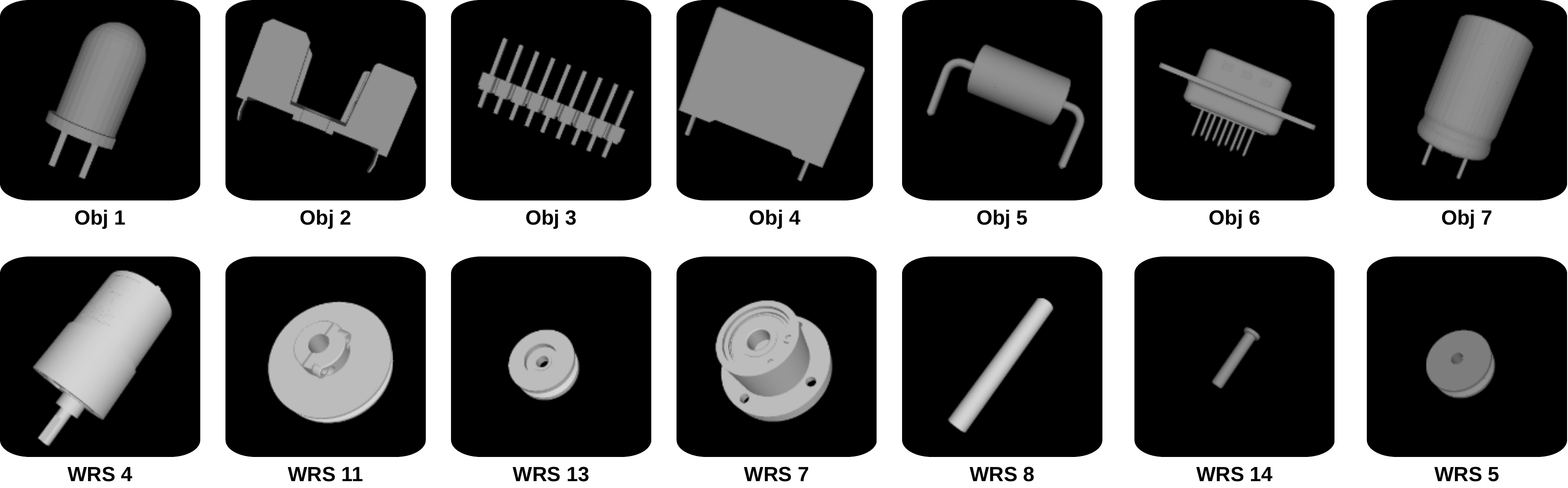}
\end{center}
   \caption{Top: The seven electronic components in the test dataset. Bottom: The seven components from the WRS dataset used for out of class tests. }
\label{fig:objects}
  \vspace{-3mm}
\end{figure*}

\section{Experiments}
For all experiments a single model is trained.
The training data consists of 1,500 CAD models from an online database of electronic components\footnote{\url{https://www.pcb-3d.com/membership_type/free/}}. Fifty objects are excluded for validation.
%

The test dataset consists of seven electrical components introduced in a different dataset\cite{hagelskjaer2022hand}.
The seven test objects are shown at the top of Fig.~\ref{fig:objects}.
%

Additionally we test the ability of the network on out of class objects. Seven industrial objects from the WRS \cite{yokokohji2019assembly} dataset is used for testing. On this dataset we show the networks ability to generalize to other objects outside of the training scope.
The objects from the WRS dataset are shown in the bottom of Fig.~\ref{fig:objects}.

Finally, results are shown for real test data. The method is compared with a state-of-the-art colorless pose estimation algorithm \cite{hagelskjaer2022parapose}. Results are not shown for zero-shot methods requiring RGB information as this is not available for the objects.

%
%
All point cloud processing was performed using the Open3D framework \cite{Zhou2018}.
The network processing was performed using PyTorch \cite{Paszke_PyTorch_An_Imperative_2019}.

\subsection{Network Training}
The network was trained on a PC environment with two NVIDIA GeForce RTX 2080 GPUs. The network was trained for 120 epochs lasting approximately six days (141 hours). For each object we generate 160 point clouds giving an epoch size of 232,000.
%
%

The network is trained with a batch size of 14, with 7 on each GPU, using the Adam optimizer \cite{kingma2014adam}, with an initial learning rate of 0.0001. We use a step scheduler with a step size of 20 and the gamma parameter set to 0.7. The loss is calculated using cross entropy with segmentation and keypoint loss weighted 0.2 and 0.8, respectively. For the keypoint loss only points belonging to the object are used. 
Group norm \cite{wu2018group} with size 32 is used as opposed to Batch Norm as a result of the small batch size.
%

The dropout is set to 40~\%, and applied as shown in Fig.~\ref{fig:network}. The dropout on the object features is used as the network should not overfit to specific parts of the object.
Additionally, up to 0.7~\% Gaussian noise is applied to the object and scene point clouds, and 10~\% position shift is applied to the object point cloud.



\subsection{Training and test performance}

\begin{figure*}[t]
    \begin{center}
        \vspace{1.5mm}
\begin{tabular}{c@{\hskip 0.04in}ccccc}
        \rotatebox[origin=lB]{90}{\ \ \ \ \ \ Obj 1} &   
        \cutpic{280}{50}{280}{50}{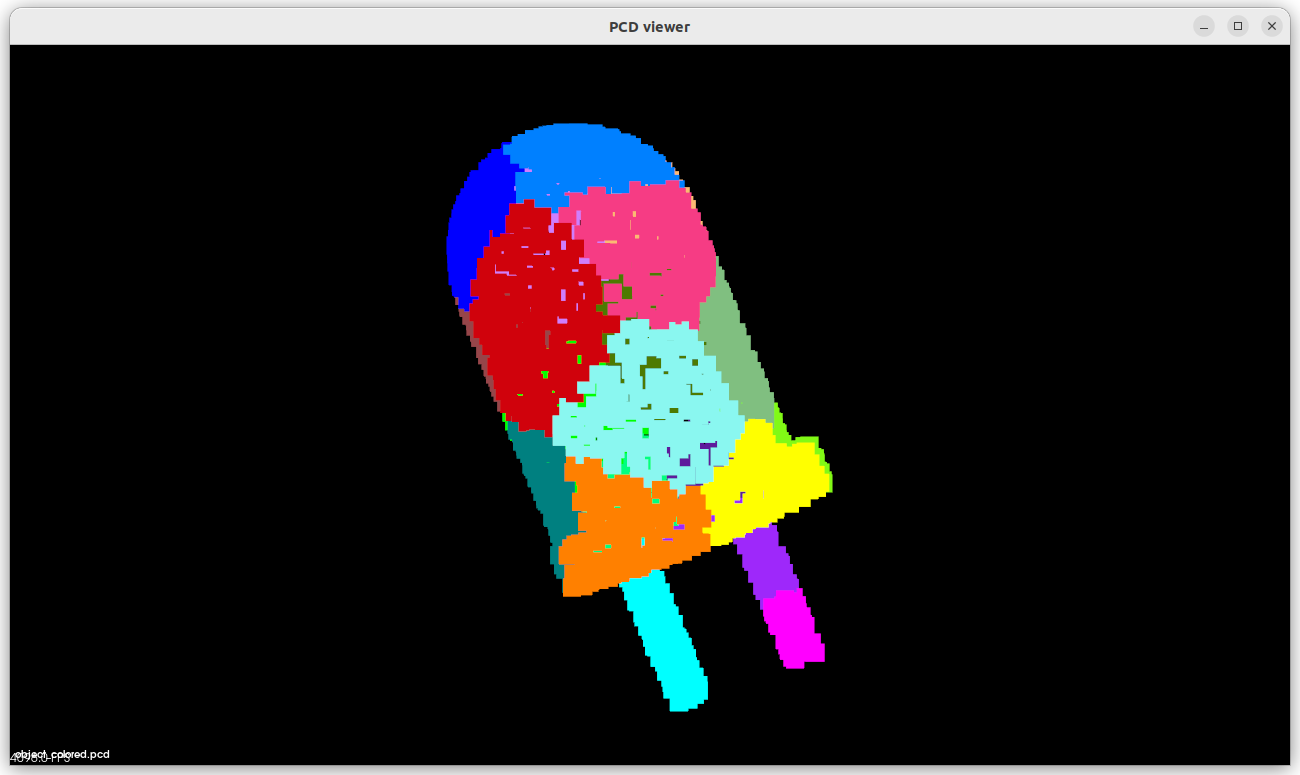} & 
        \cutpic{320}{100}{370}{120}{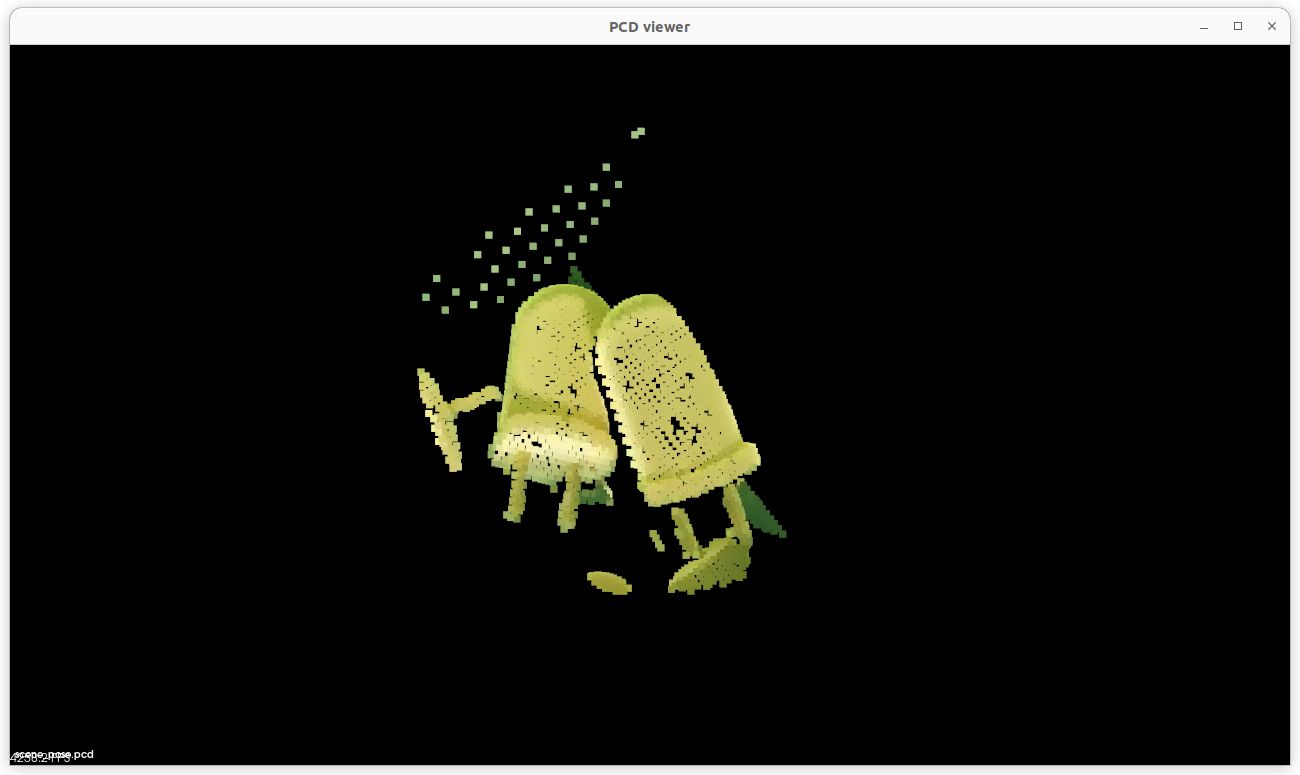} & 
        \cutpic{320}{100}{370}{120}{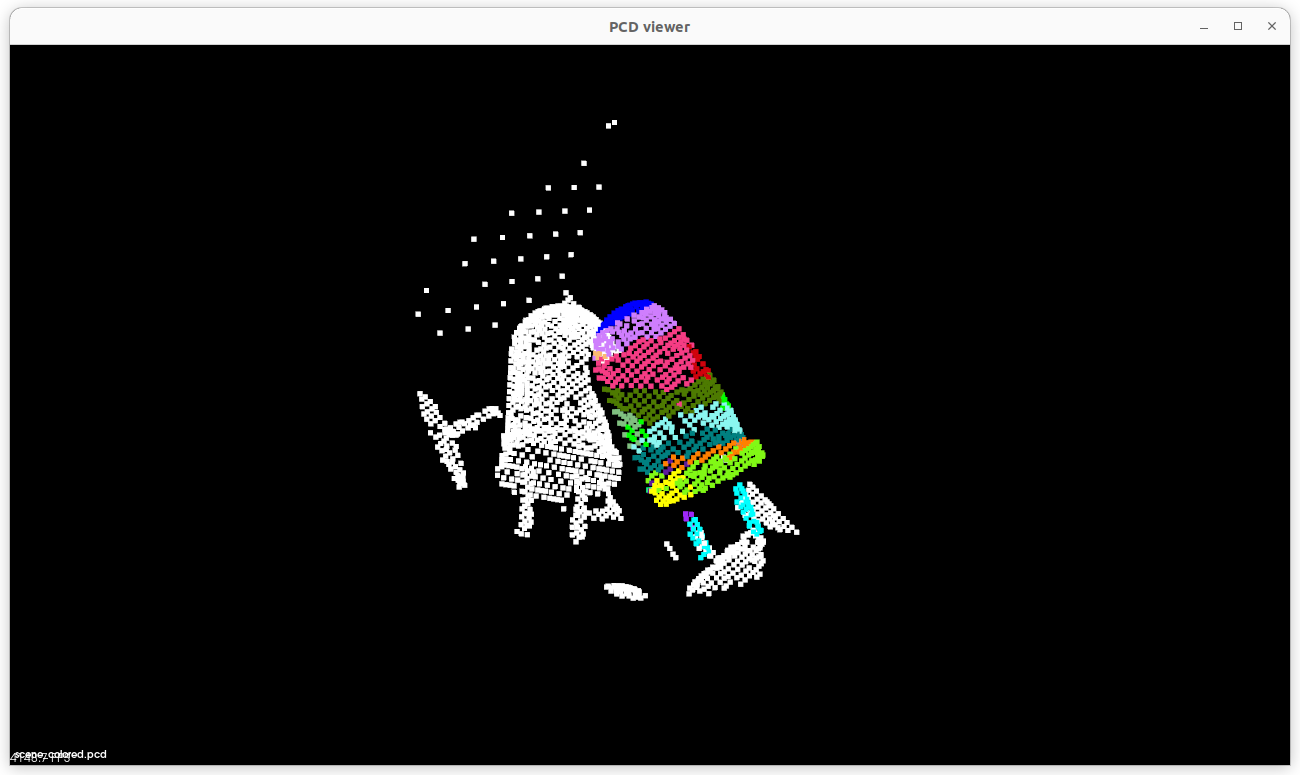} & 
        \cutpic{320}{100}{370}{120}{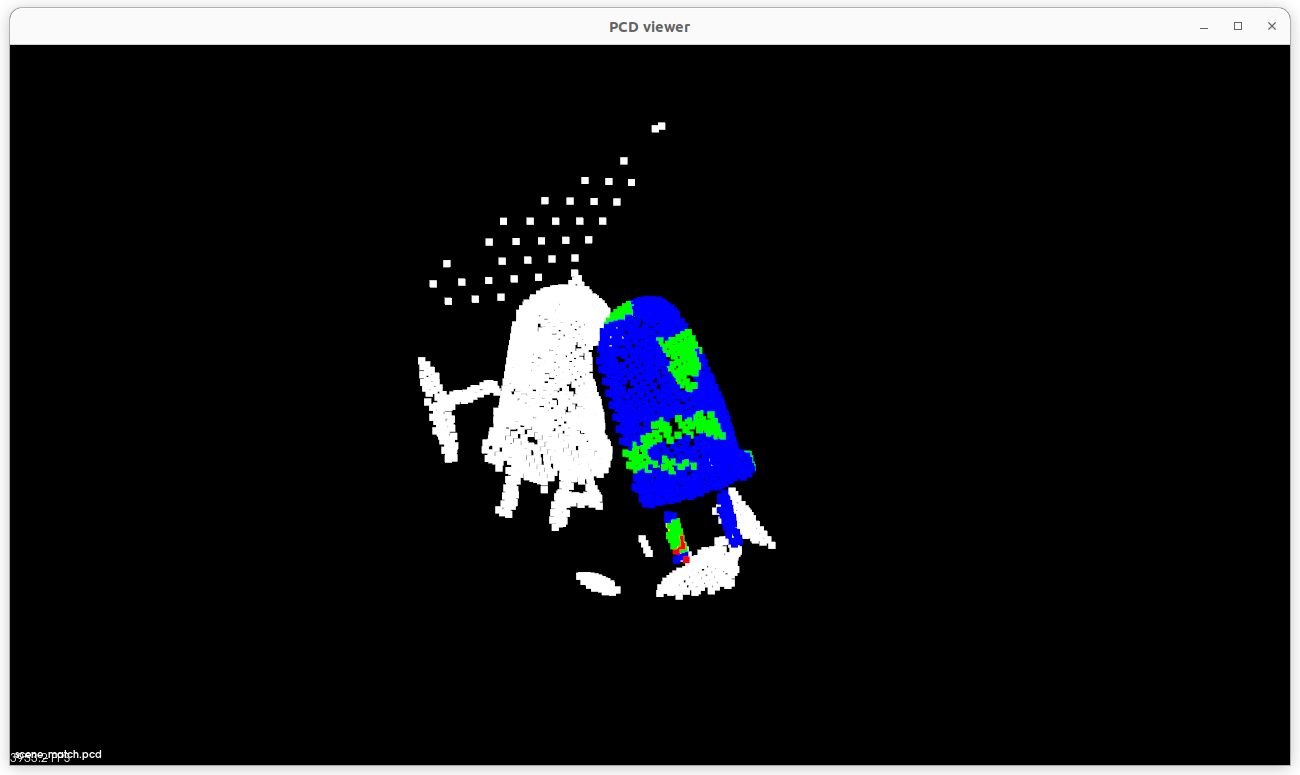} & 
        \cutpic{320}{100}{370}{120}{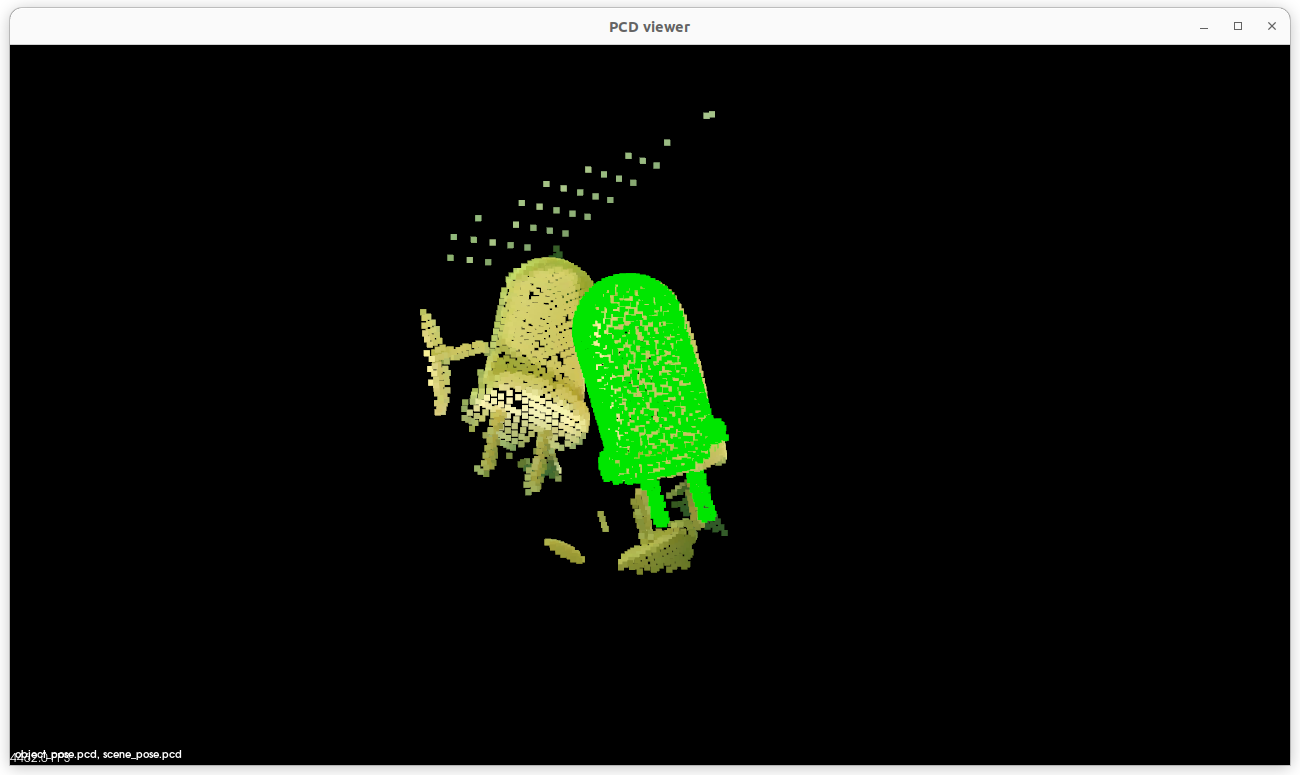} \\
        \rotatebox[origin=lB]{90}{\ \ \ \ \ \ Obj 4} & 
        \cutpic{280}{50}{280}{50}{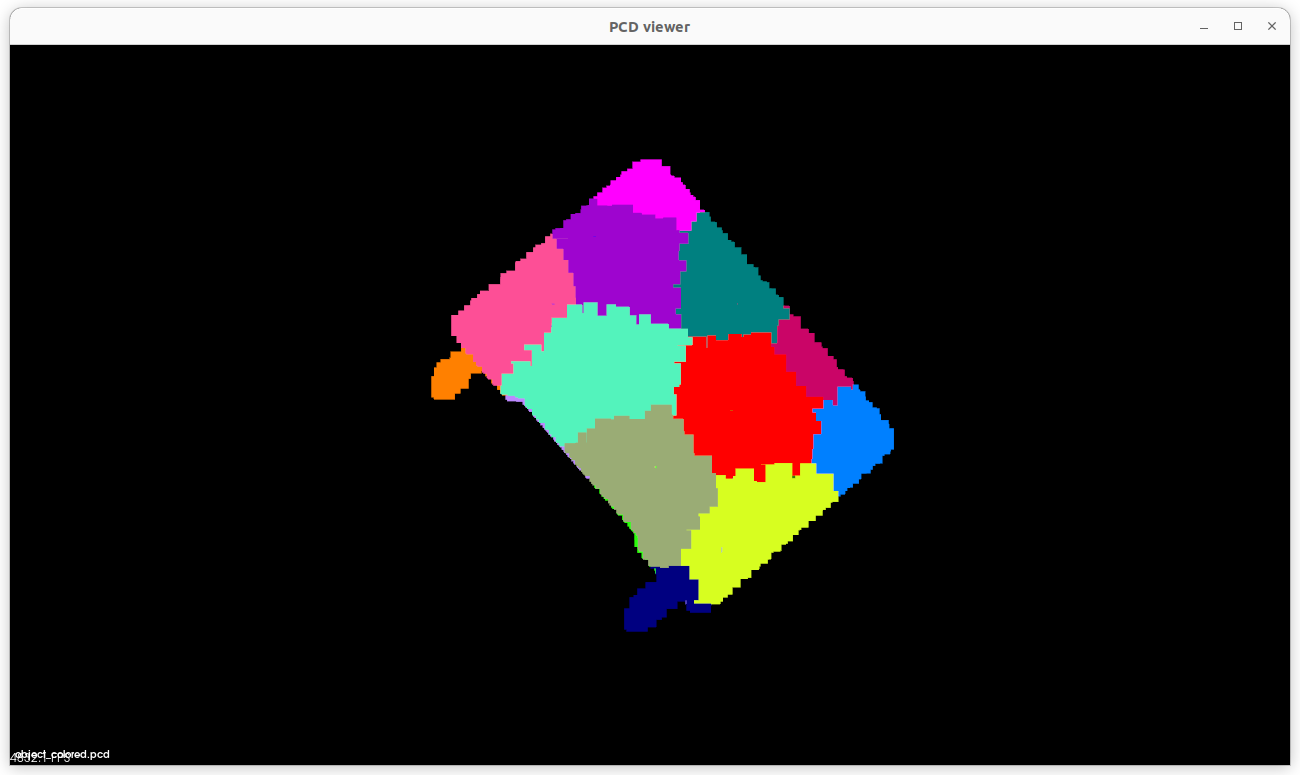} & 
        \cutpic{280}{50}{280}{50}{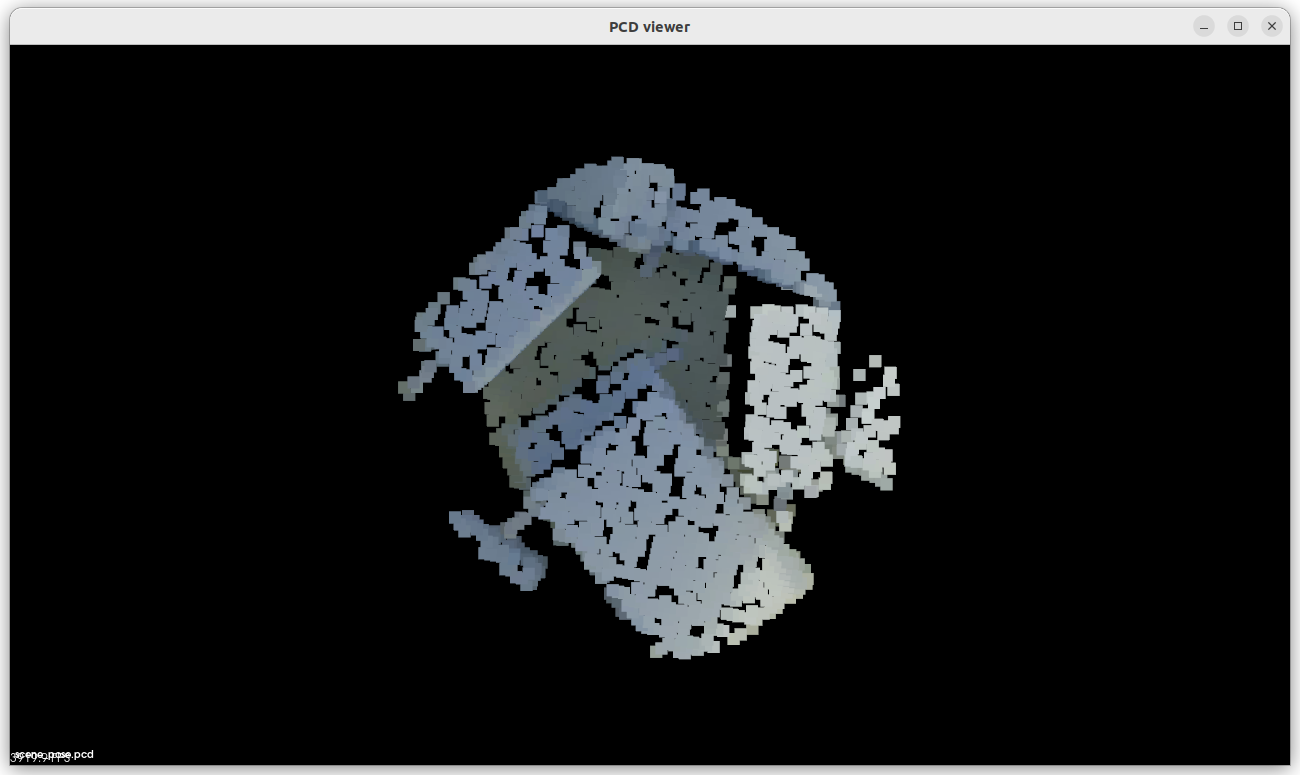} & 
        \cutpic{280}{50}{280}{50}{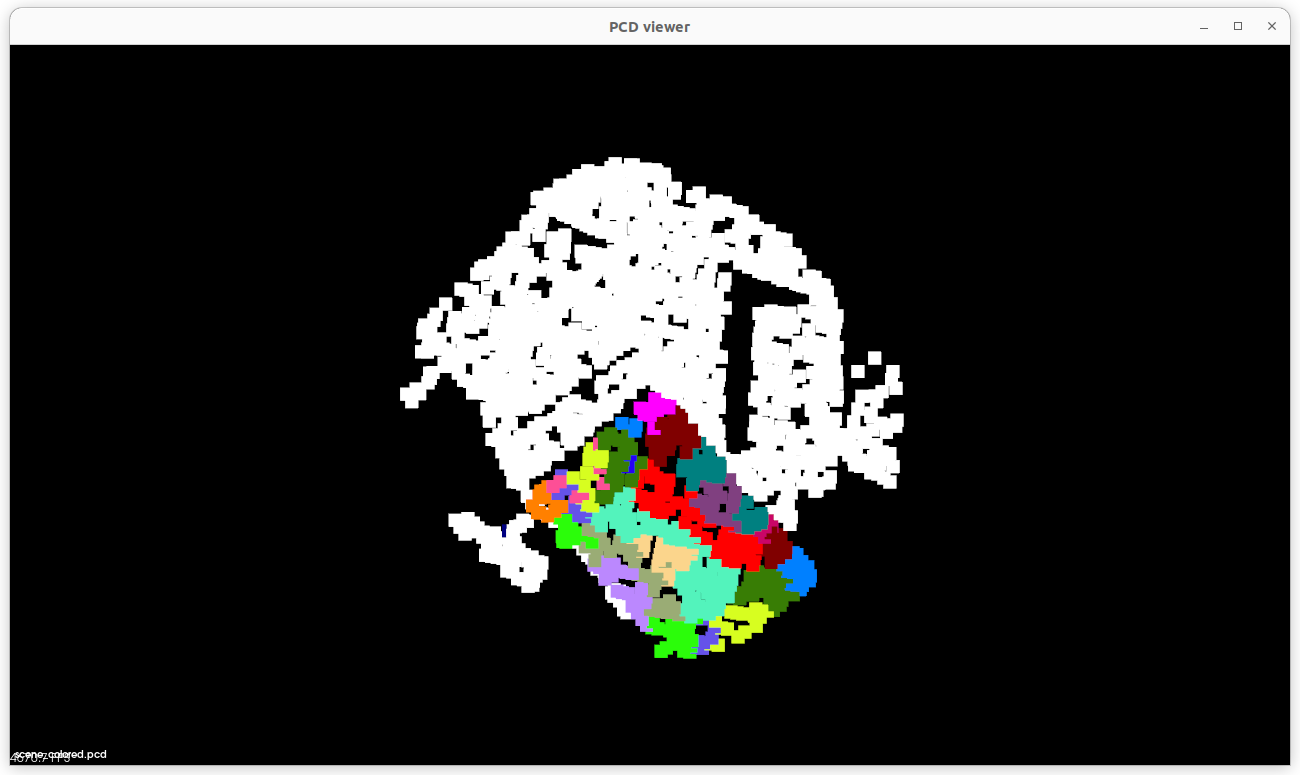} & 
        \cutpic{280}{50}{280}{50}{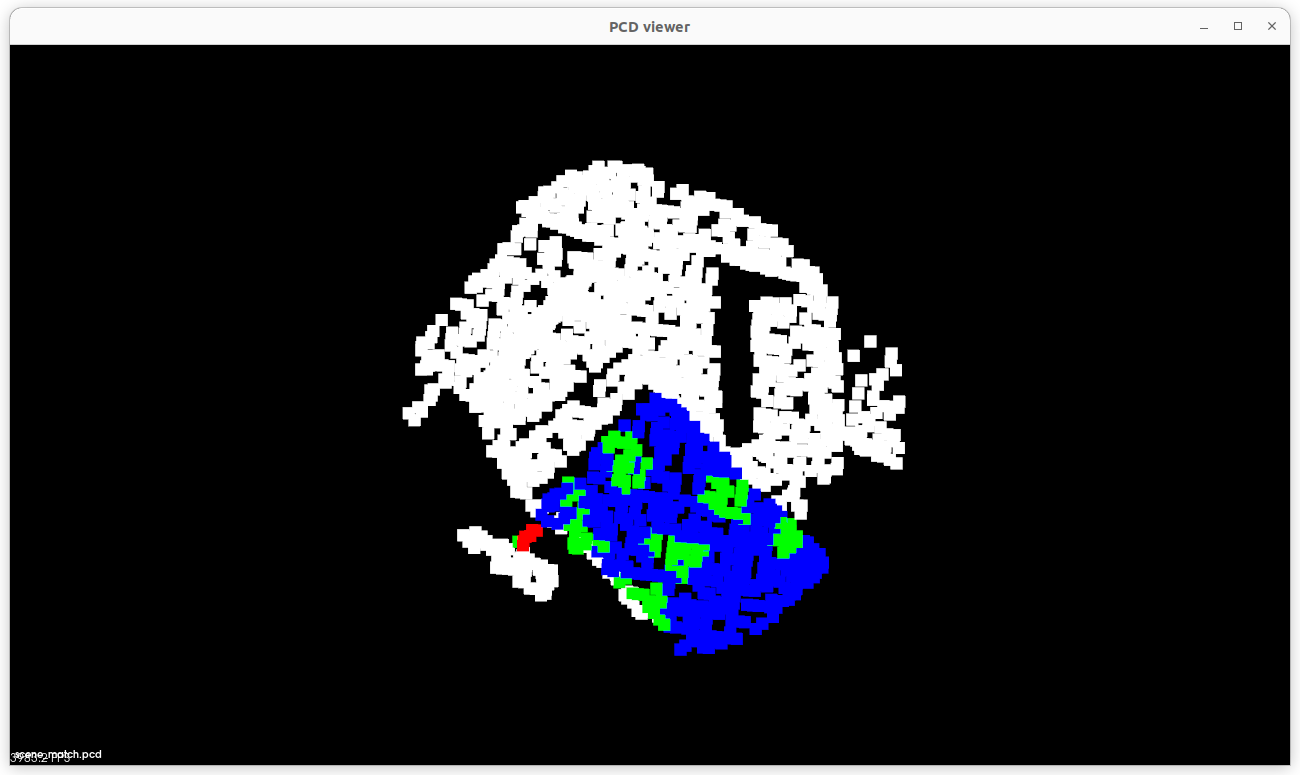} & 
        \cutpic{280}{50}{280}{50}{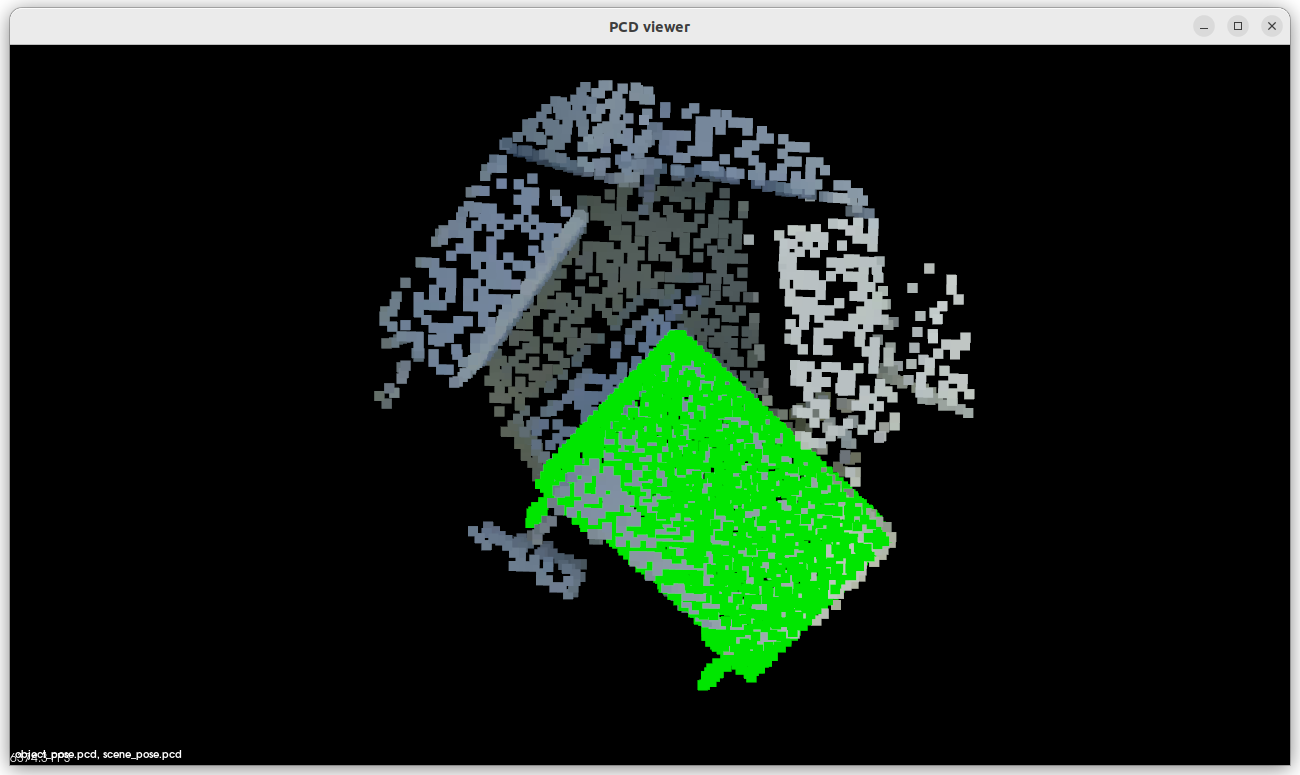} \\
        \rotatebox[origin=lB]{90}{\ \ \ \ \ \ WRS 7} & 
        \cutpic{280}{50}{280}{50}{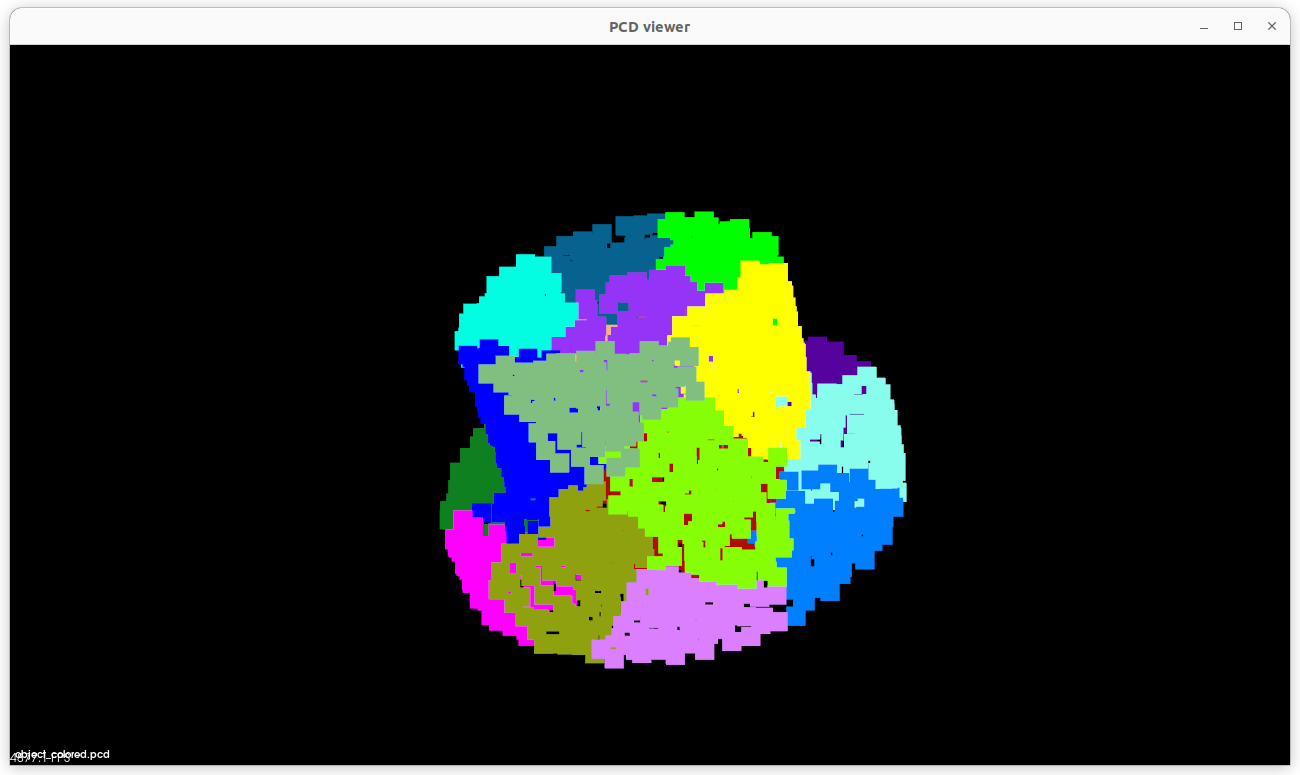} & 
        \cutpic{280}{50}{280}{50}{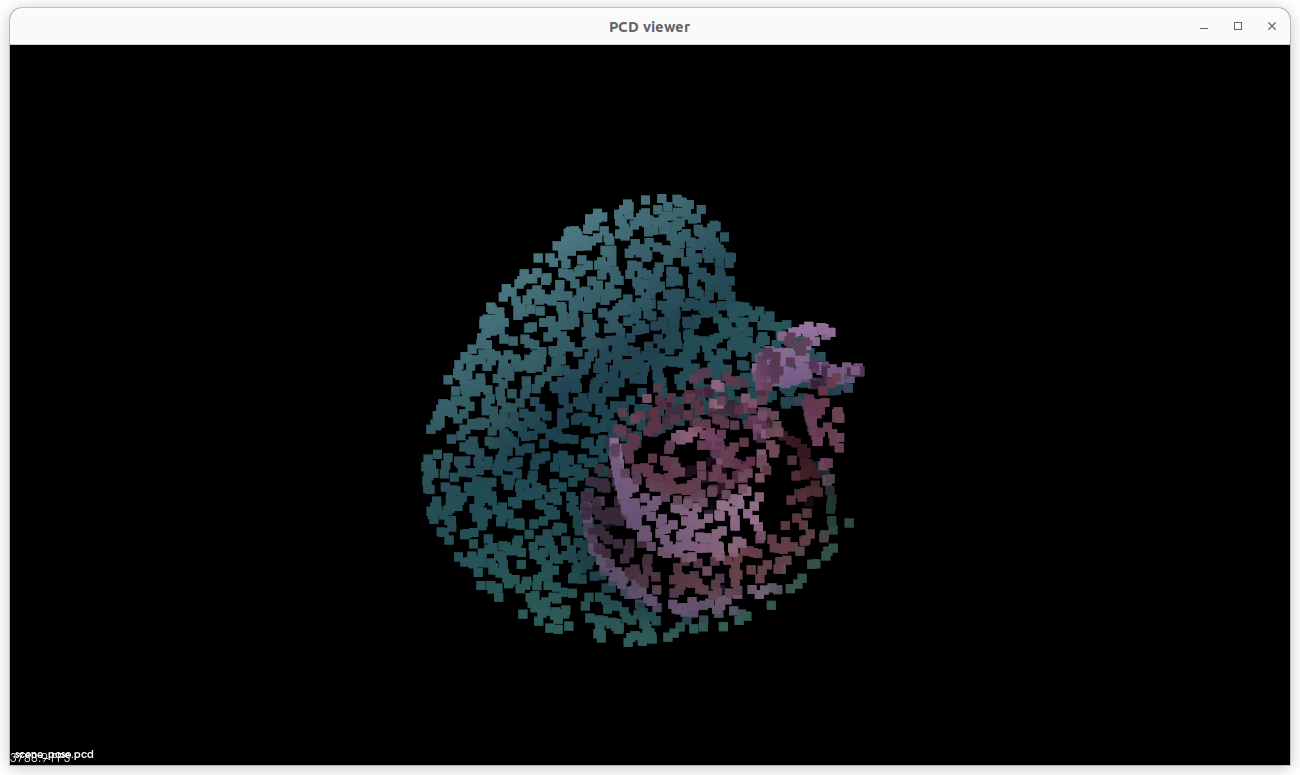} & 
        \cutpic{280}{50}{280}{50}{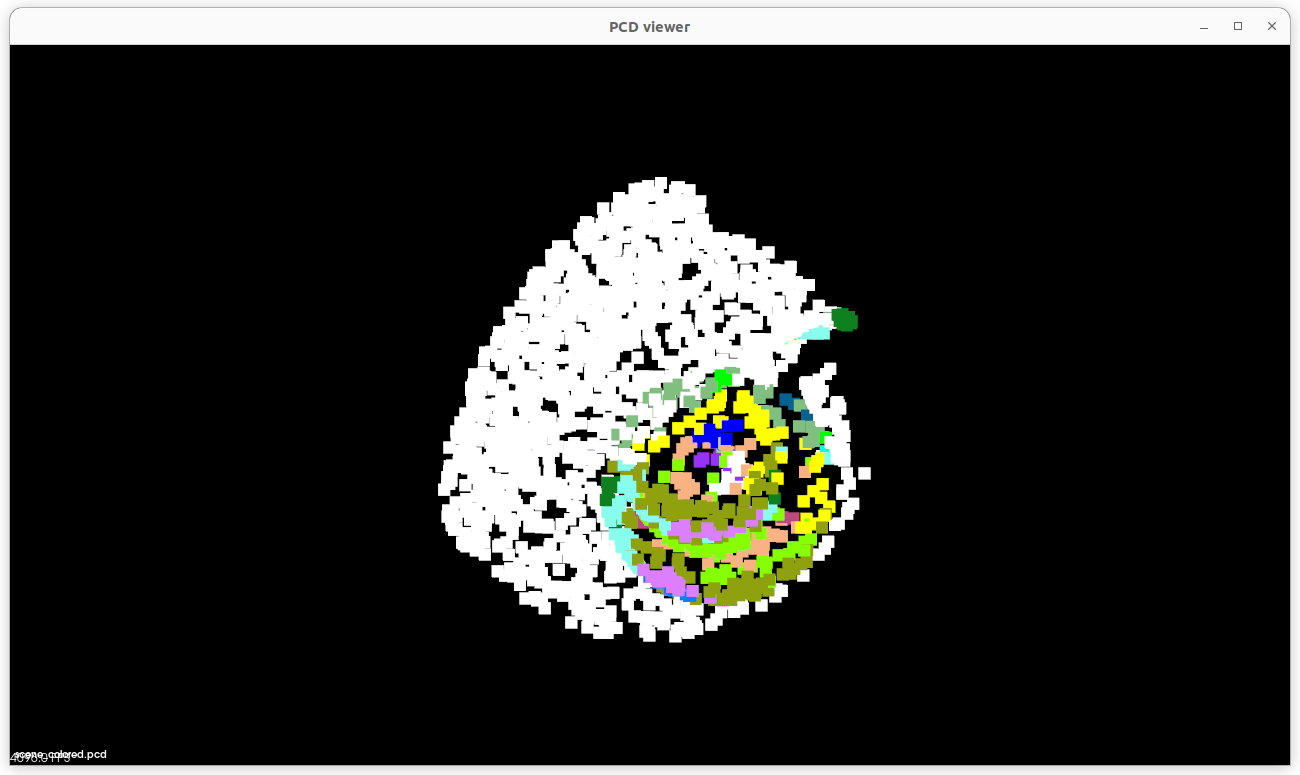} & 
        \cutpic{280}{50}{280}{50}{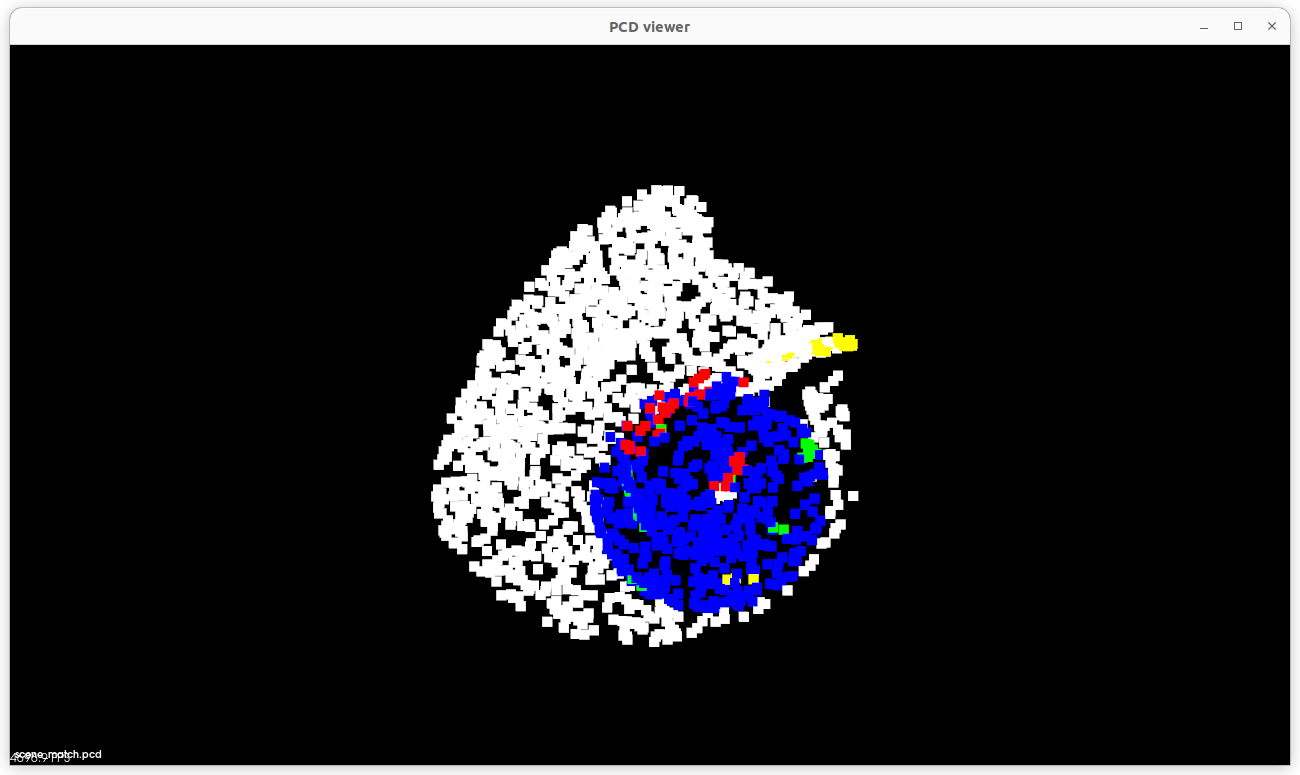} & 
        \cutpic{280}{50}{280}{50}{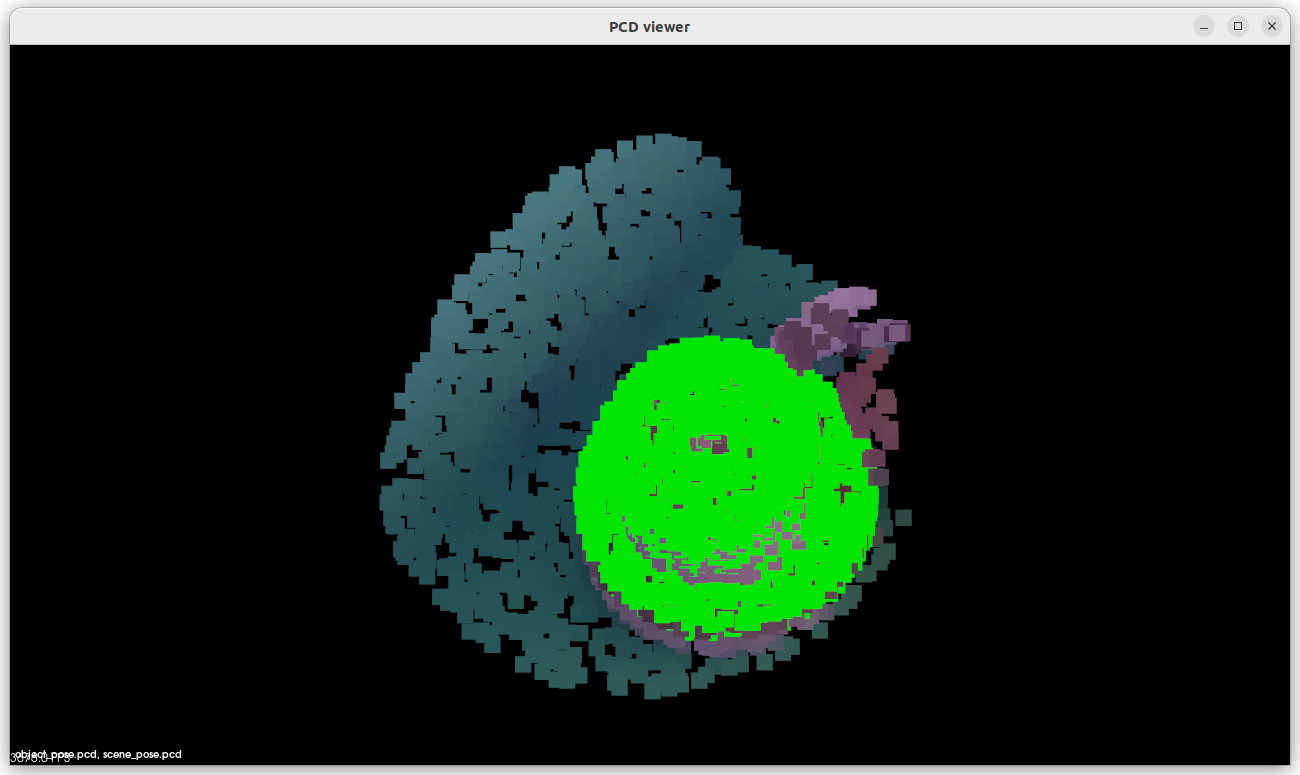} \\
        \rotatebox[origin=lB]{90}{\ \ \ \ \ \ WRS 4} & 
        \cutpic{280}{50}{280}{50}{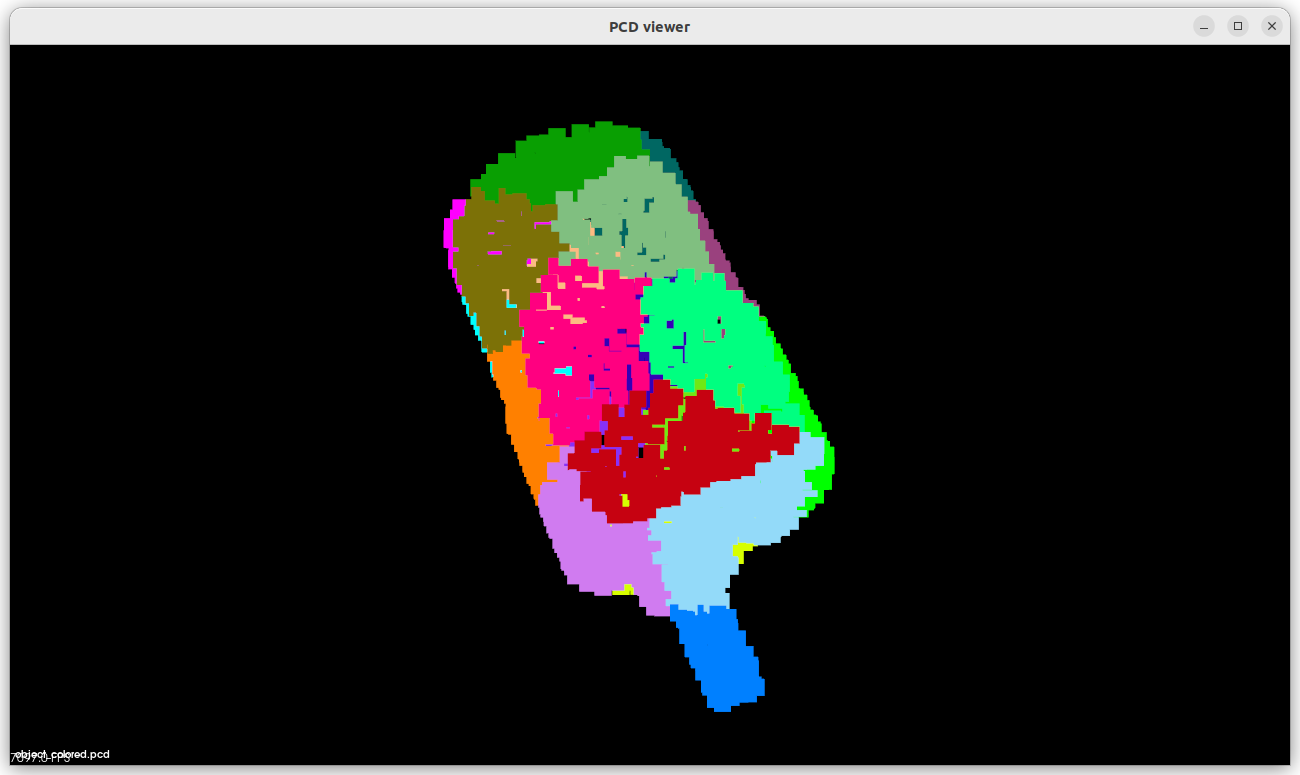} & 
        \cutpic{280}{50}{280}{50}{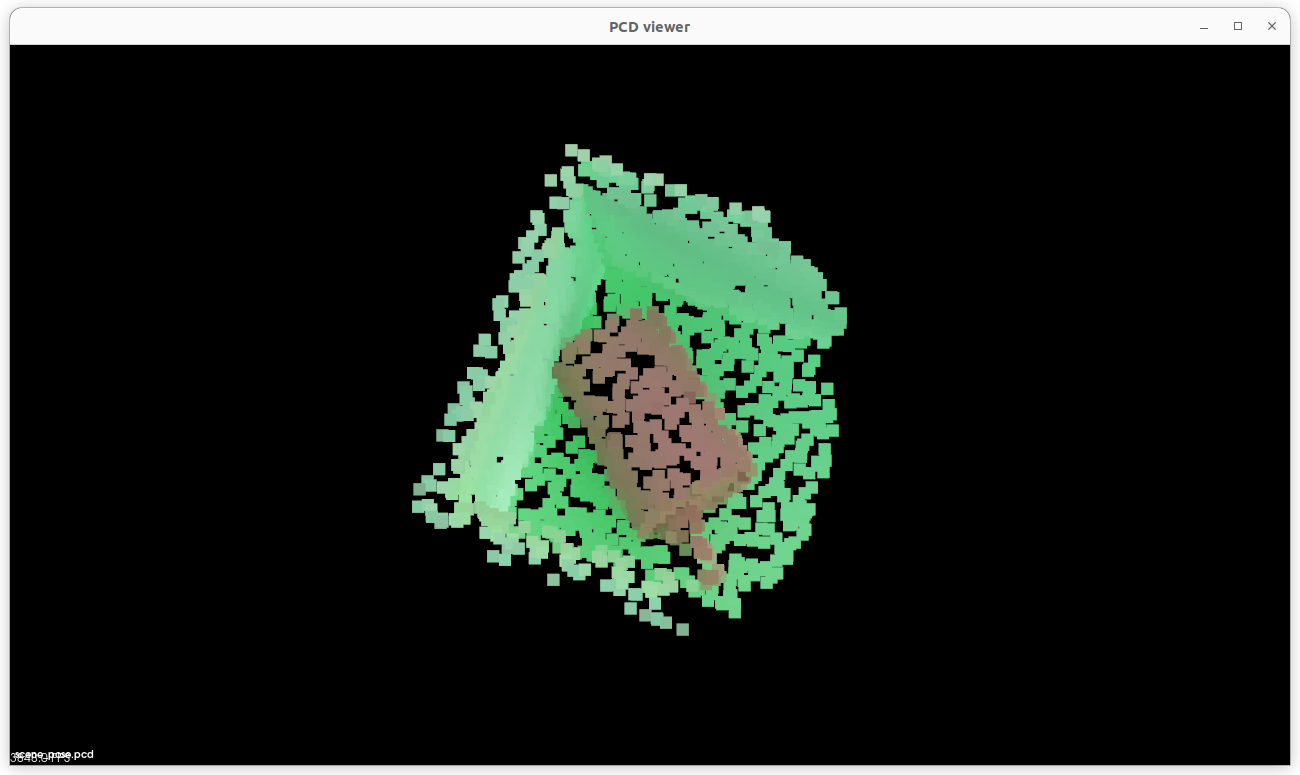} & 
        \cutpic{280}{50}{280}{50}{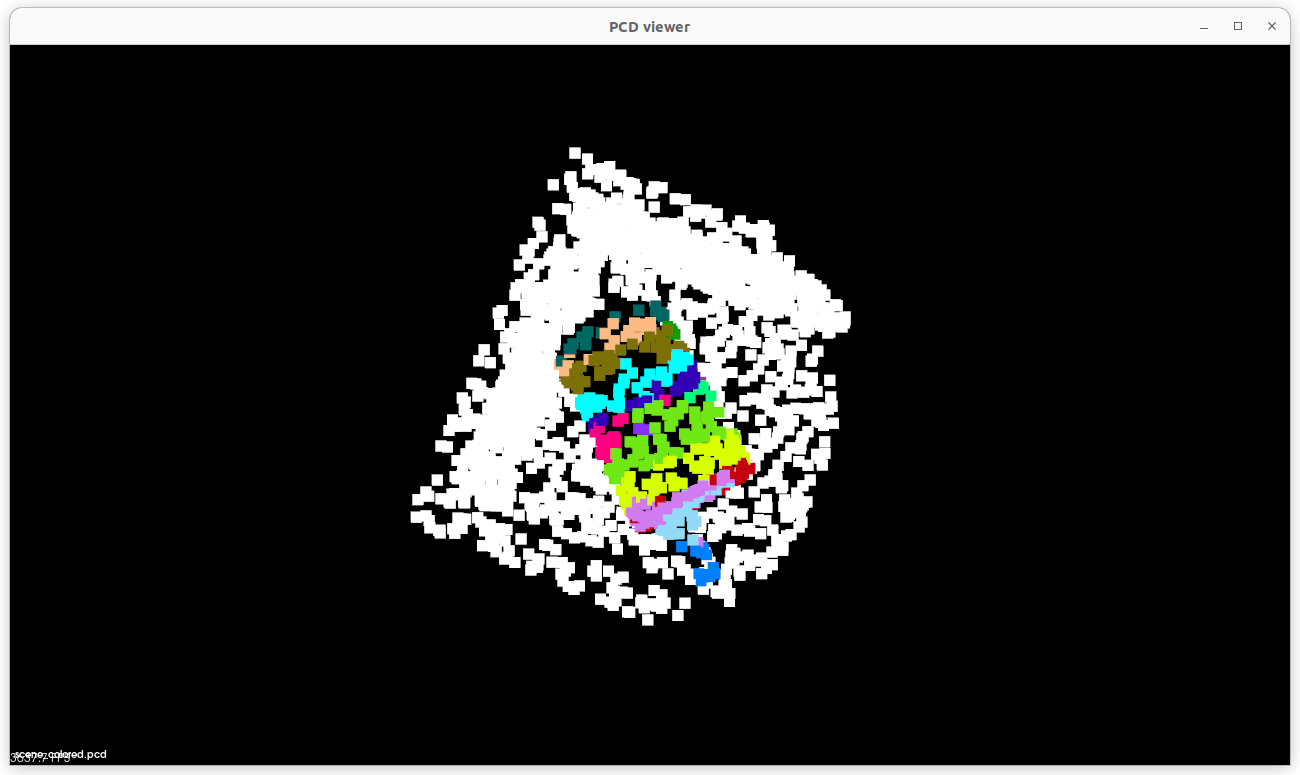} & 
        \cutpic{280}{50}{280}{50}{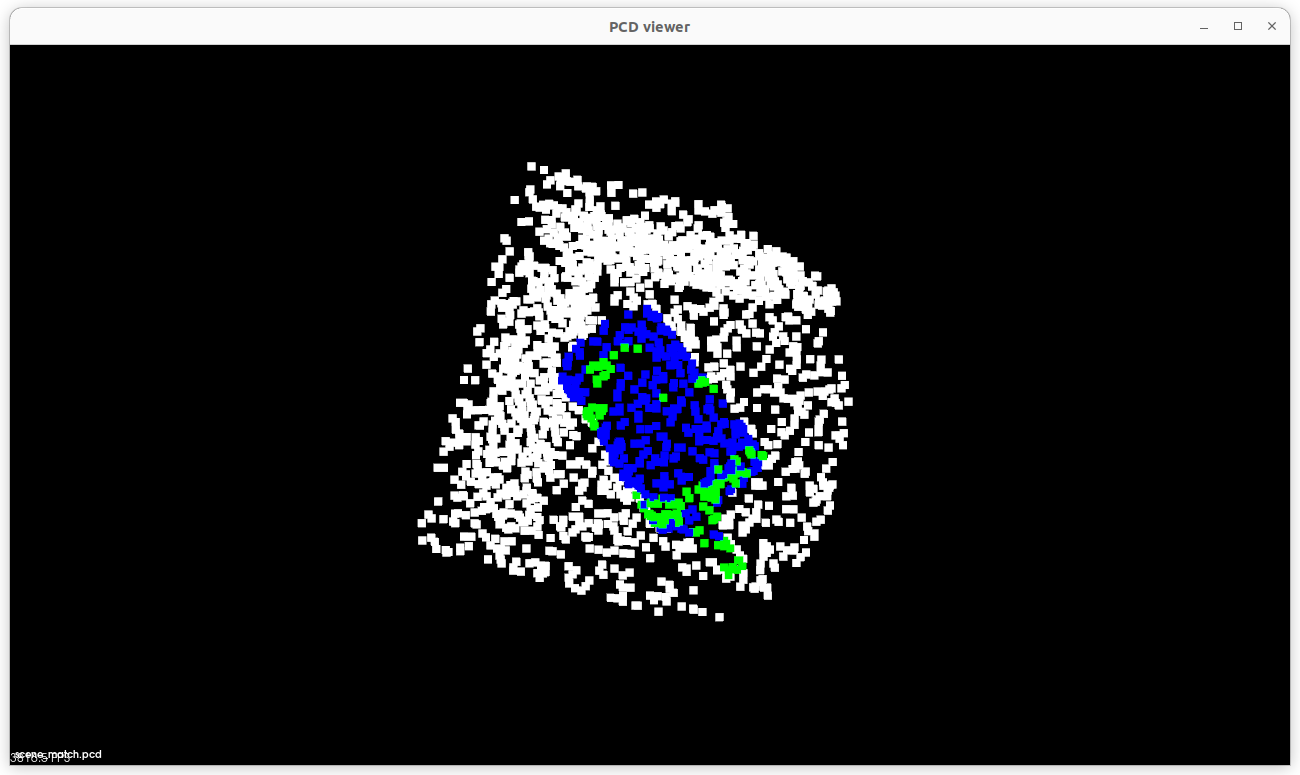} & 
        \cutpic{280}{55}{280}{40}{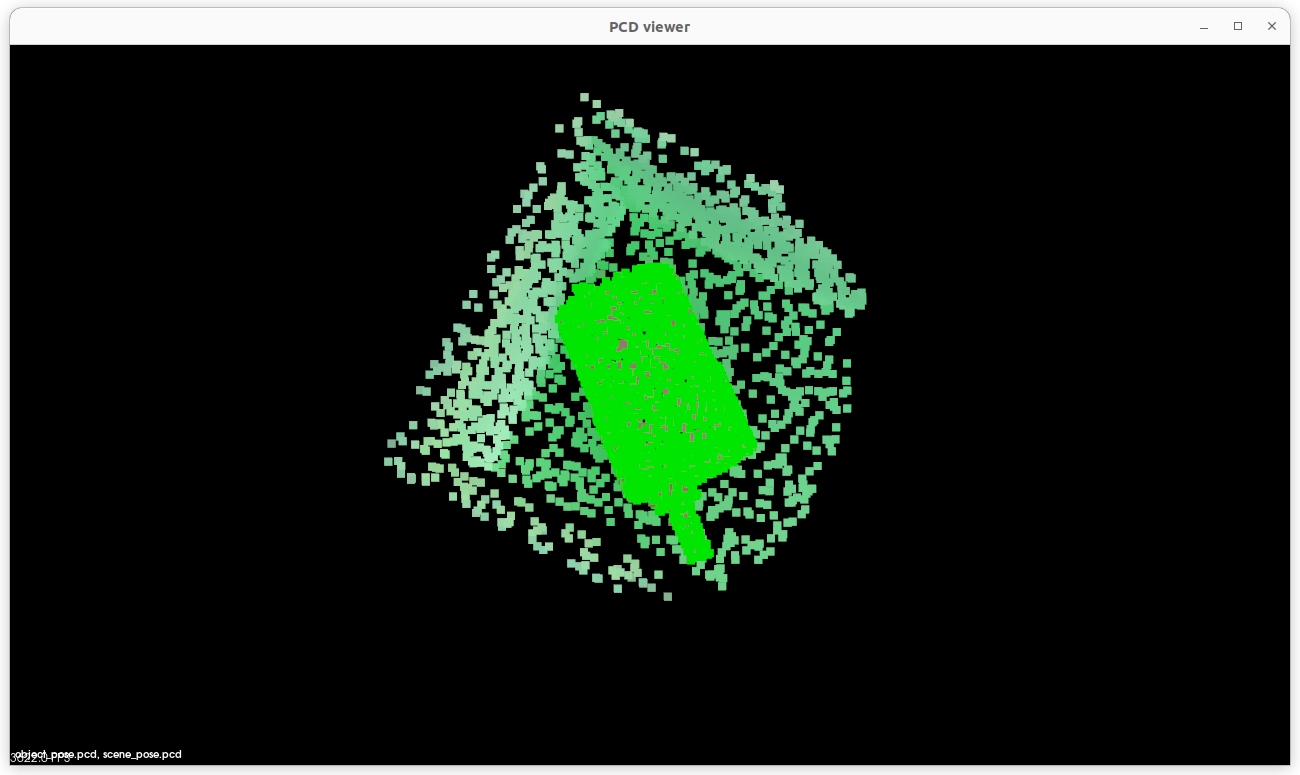} \\
       & OBJECT & SCENE & PREDICTION & ACCURACY & POSE ESTIMATION \\
       
    \end{tabular}
       
    \end{center}
       \caption{Visualization of network output and resulting pose estimation using the RANSAC from Open3D \cite{Zhou2018}. 
       The prediction accuracy is shown as follows: "White" represents correctly predicted background, "red" false negatives and "yellow" false positives. "Blue" is correctly predicted segmentation of the object, but wrong keypoints and "green" is both correct segmentation and keypoint prediction. It is seen that segmentation prediction is very high, and while the keypoint accuracy is not very high the pose estimation are correct. 
       }
    \label{fig:pe}
\end{figure*}

The performance for the trained networks is show in Tab.~\ref{tab:train}. Performance is shown for both the loss, segmentation accuracy, and keypoint accuracy. We present the performance on the training, validation and test set. The network does not appear to over-fit to the training data, and actually shows better performance on the validation set. On the test set the performance is also comparable to the training set.
%
As the objects are symmetric to varying levels, the keypoint accuracy despite being low, still gives good pose estimations. This is seen in Fig.~\ref{fig:example}, where the symmetry of the object results in striped matching of keypoints. However, these matches are still very useful for the pose estimation. This is further shown in Fig.~\ref{fig:pe}. 




\begin{table}
\begin{center}
\caption{The loss and accuracy of the network. The parentheses indicate using 10\% of the training data. }
\begin{tabular}{|l|c|c|c|}
\hline
Split & Training & Validation & Test \\
\hline\hline
Loss      & 1.59 (1.63) & 1.33 (1.40) & 1.56 (1.65) \\ 
Seg. Acc  & 0.98 (0.98) & 0.96 (0.94) & 0.95 (0.94) \\ 
Key. Acc  & 0.27 (0.28) & 0.31 (0.29) & 0.27 (0.24) \\ 
\hline
\end{tabular}
\label{tab:train}
 \vspace{-6mm}
\end{center}
\end{table}



To analyze the network, performance for each component is shown in Tab.~\ref{tab:components}. The two objects with the smallest loss is "3" and "5". These two object are both very similar to objects in the training data. The most challenging object is "2". The split between the two parts of the object makes it very dissimilar to the training data. The other components perform very well, especially for the segmentation task.

\noindent \textbf{Training Samples:} An experiment was performed to test the influence of the number of training objects. Ten percent of the objects were randomly sampled to be used for training. The number of iterations per epoch was kept the same as in the original training. 
While the results in Tab.~\ref{tab:train} show a slight decrease in accuracy, it is seen that even by 10~\% of the training data the model generalizes well. This is possibly a result of the simplified domain problem and indicates that the amount of real training data necessary is not overwhelmingly large.


\begin{table}
\begin{center}
\caption{Loss and accuracy of each of the individual test objects.}
\begin{tabular}{|l|c|c|c|c|c|c|c|}
\hline
Object   & 1    & 2    & 3    & 4    & 5    & 6    & 7    \\ \hline 
\hline
Loss     & 1.50 & 2.03 & 1.12 & 1.53 & 1.42 & 1.84 & 1.53 \\ 
Seg. Acc & 0.99 & 0.81 & 0.98 & 0.98 & 0.95 & 0.97 & 0.99 \\ 
Key. Acc & 0.24 & 0.16 & 0.43 & 0.29 & 0.31 & 0.22 & 0.22 \\ \hline
\end{tabular}
\label{tab:components}
\end{center}
\end{table}

\begin{table}
\begin{center}
	\setlength\tabcolsep{5.0pt}
\caption{Pose estimation recall for each of the test components. The "\%" sign indicates the amount of noise added to the scene point cloud.}
    \begin{tabular}{|l|c|c|c|c|c|c|c|c|}
    \hline
    Object  & 1  & 2  & 3  & 4  & 5  & 6  & 7 & Avg.  \\ \hline
    \hline
	    Ours                & 0.95 & 0.82 & 0.99 & 0.77 & 0.91 & 0.90 & 0.92 & 0.89 \\
	    FPFH             & 0.55 & 0.62 & 0.61 & 0.30 & 0.36 & 0.67 & 0.51 & 0.52 \\
    \hline
	    Ours 1\%  & 0.93 & 0.79 & 0.99 & 0.74 & 0.86 & 0.88 & 0.90 & 0.87 \\
	    FPFH 1\%  & 0.43 & 0.57 & 0.50 & 0.25 & 0.31 & 0.66 & 0.40 & 0.44 \\
    \hline
	    Ours 5\% & 0.55 & 0.60 & 0.91 & 0.62 & 0.83 & 0.82 & 0.53 & 0.69 \\
	    FPFH 5\% & 0.30 & 0.54 & 0.45 & 0.20 & 0.43 & 0.56 & 0.25 & 0.39 \\
    \hline
    \end{tabular}
\label{tab:classic}
\end{center}
  \vspace{-6mm}
\end{table}

\subsection{Pose Estimation Performance}

To test the pose estimation accuracy of the developed method we compare it with the classic pose estimation method FPFH \cite{rusu2009fast}. For both methods, RANSAC is used \cite{fischler1981random}. Additionally, we test our method using the GPU based RANSAC. The performance is measured by the ADI score presented in \cite{hinterstoisser2012model}, as it is well suited for the symmetric objects in the dataset. The classic method obtains a $0.57\%$ mean accuracy whereas our method obtains a $0.95\%$ mean accuracy. Using the GPU based RANSAC the mean accuracy is $0.86\%$. A slightly lower performance. Our method thus vastly outperforms the classic pose estimation method.

As in \cite{buch2017rotational} Gaussian noise is added to the points clouds to further test the robustness of the system. 
The Gaussian noise is tested at 1~\% and 5~\% of the longest axis of the object. 
The results are shown in Tab.~\ref{tab:classic}. It can be seen that our method outperforms the classic method for all objects. 
When adding noise the difference becomes even more pronounced. However, for the $5\%$ noise the performance also drops significantly for our method.

\begin{table}
\begin{center}
	\setlength\tabcolsep{4.0pt}
\caption{Pose estimation recall on the out-of-class dataset. The objects are numbered according to \cite{yokokohji2019assembly}.
The "\%" sign indicates the amount of noise added to the scene point cloud.} 
\begin{tabular}{|l|c|c|c|c|c|c|c|c|}
\hline
	Number   & 4   & 11    & 13    & 7    & 8    & 14    & 5  & Avg.   \\ 
	Type    & {\scriptsize Motor} & {\scriptsize Pulley} & {\scriptsize Idler} & {\scriptsize Bear.} & {\scriptsize Shaft} & {\scriptsize Screw} & {\scriptsize Pulley} &  \\ 
\hline
\hline
	Ours & 1.00 & 0.99 & 0.90 & 0.91 & 0.93 & 0.93 & 0.96 & 0.95 \\
	GPU  & 1.00 & 0.99 & 0.98 & 0.96 & 0.90 & 0.97 & 0.98 & 0.97 \\
	FPFH & 0.98 & 0.93 & 0.78 & 0.78 & 0.67 & 0.37 & 0.90 & 0.77 \\ \hline

	Ours  1\% & 1.00 & 0.99 & 0.89 & 0.94 & 0.93 & 0.92 & 0.96 & 0.95\\
	GPU 1\% & 1.00 & 0.99 & 0.97 & 0.97 & 0.89 & 0.96 & 0.98 & 0.97 \\
	FPFH 1\% & 0.95 & 0.91 & 0.75 & 0.79 & 0.65 & 0.30 & 0.89 & 0.95 \\ \hline

	Ours 5\% & 0.97 & 0.98 & 0.87 & 0.94 & 0.82 & 0.73 & 0.95 & 0.89 \\
	GPU 5\% & 0.97 & 0.97 & 0.88 & 0.95 & 0.67 & 0.70 & 0.92 & 0.87 \\
	FPFH 5\% & 0.67 & 0.89 & 0.74 & 0.82 & 0.57 & 0.28 & 0.80 & 0.68 \\ \hline

\end{tabular}
\label{tab:wrs}
\end{center}
 \vspace{-4mm}
\end{table}

\textbf{Testing out of class:}
The out-of-class dataset consists of industrial objects, such as motors and pulleys. The objects were used for the WRS assembly challenge held in 2018 \cite{yokokohji2019assembly}. The objects were chosen as they represent an industrial challenge, and because of the variety. To further test the robustness of the system, varying levels of noise is added to the points clouds. 

The results of the network performance and pose estimation is shown in Tab.~\ref{tab:wrs}. It is seen that our developed method obtains very high recall on these objects that appear quite different compared with the training data.  When adding noise the difference becomes even more pronounced. The ability to correctly compute features for both in and out-of-class objects is shown in Fig.~\ref{fig:pe}. The GPU based method actually shows higher recall than the comparison method. However, these is a noticeable drop for WRS 8 (Shaft) at 5~\% noise.

\begin{figure*}[hthp]
    \vspace{1.5mm}
    \begin{center}
        \begin{tabular}{c@{\hskip 0.04in}ccccc}
           \rotatebox[origin=lB]{90}{\ \ \ \ \ \ SCENE} & 
           \cutpic{600}{80}{500}{160}{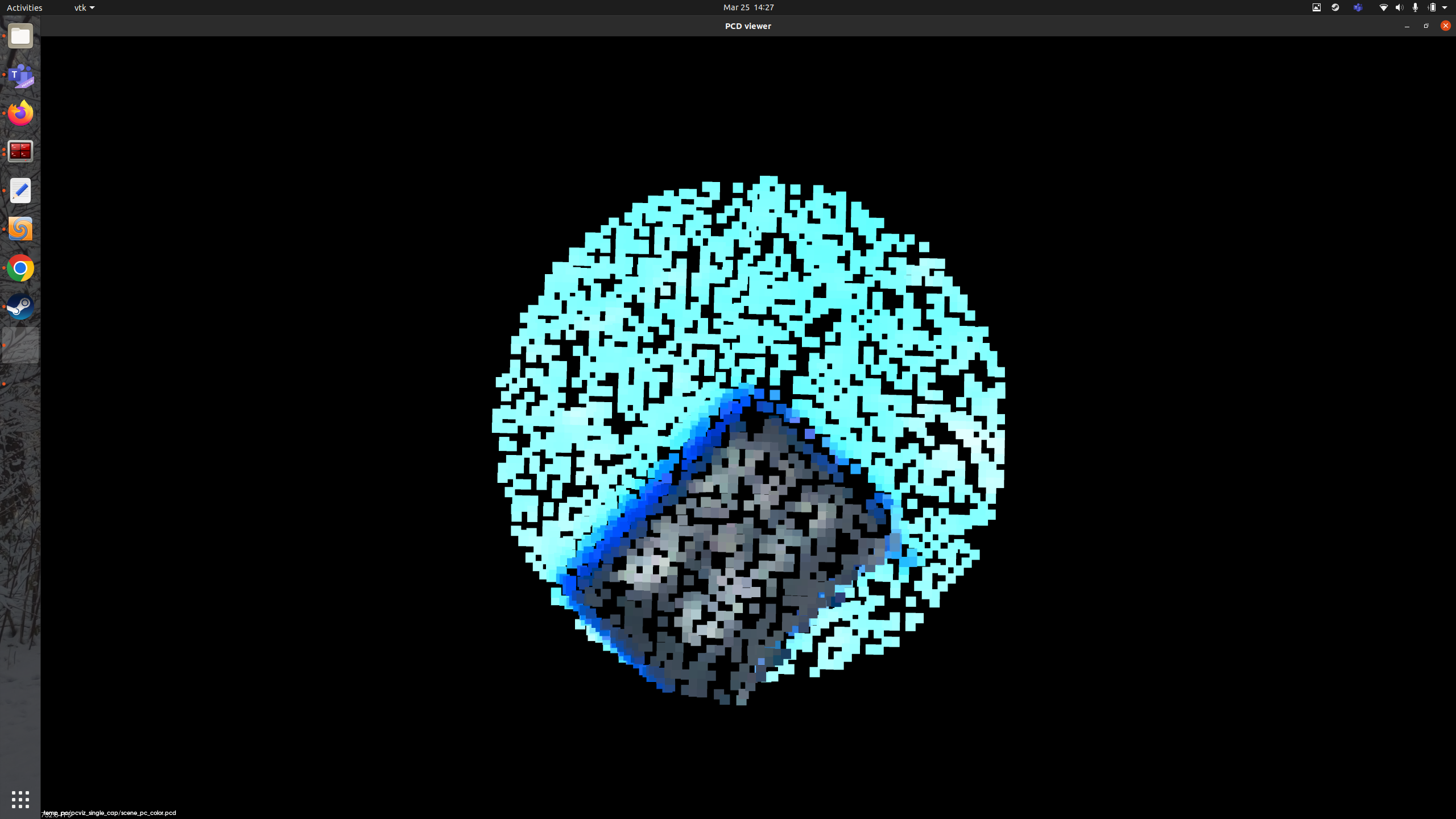} & 
            \cutpic{600}{80}{500}{160}{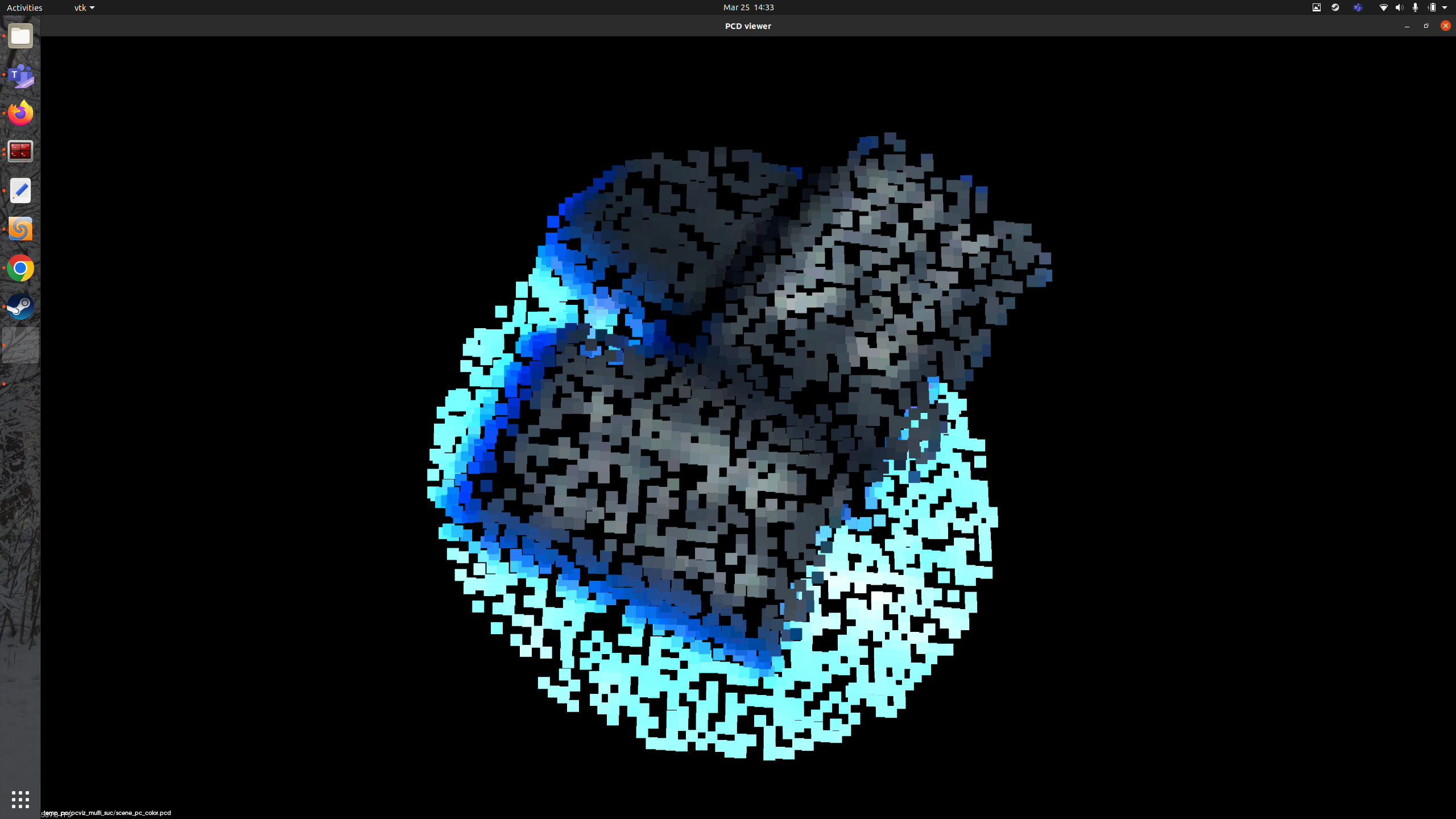} & 
            \cutpic{600}{80}{500}{160}{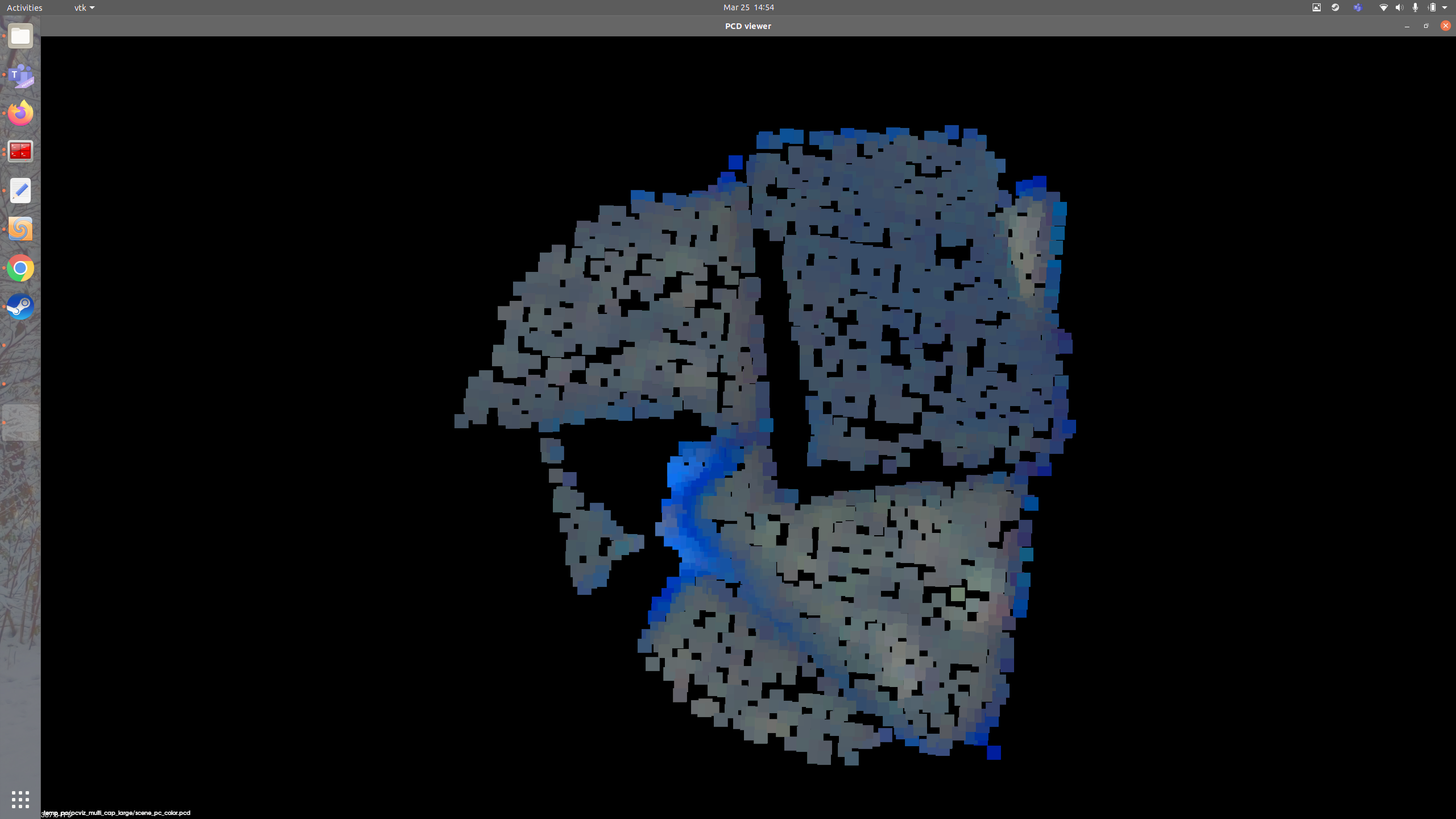} &
            \cutpic{600}{120}{500}{120}{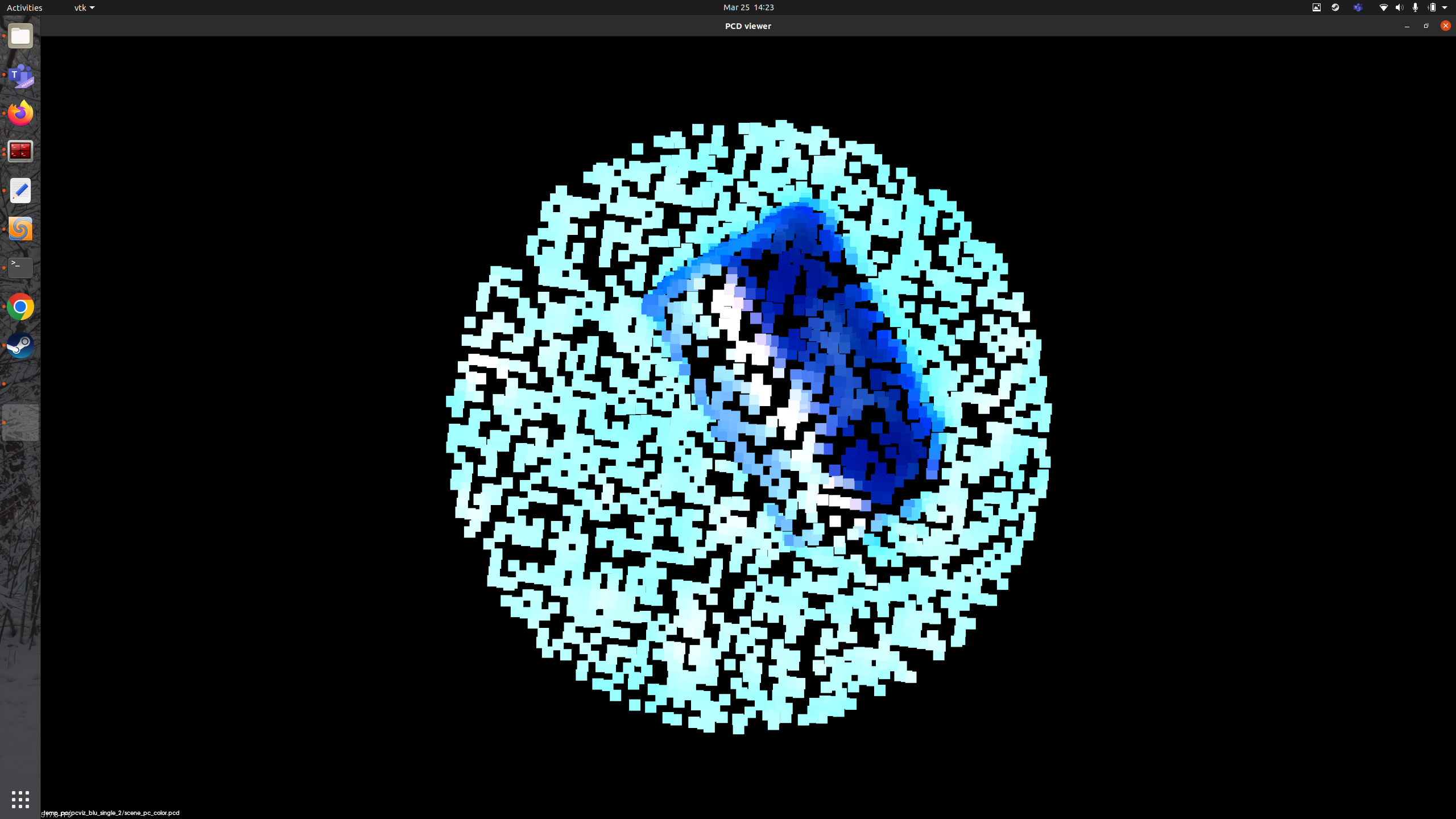} & 
            \cutpic{600}{120}{500}{120}{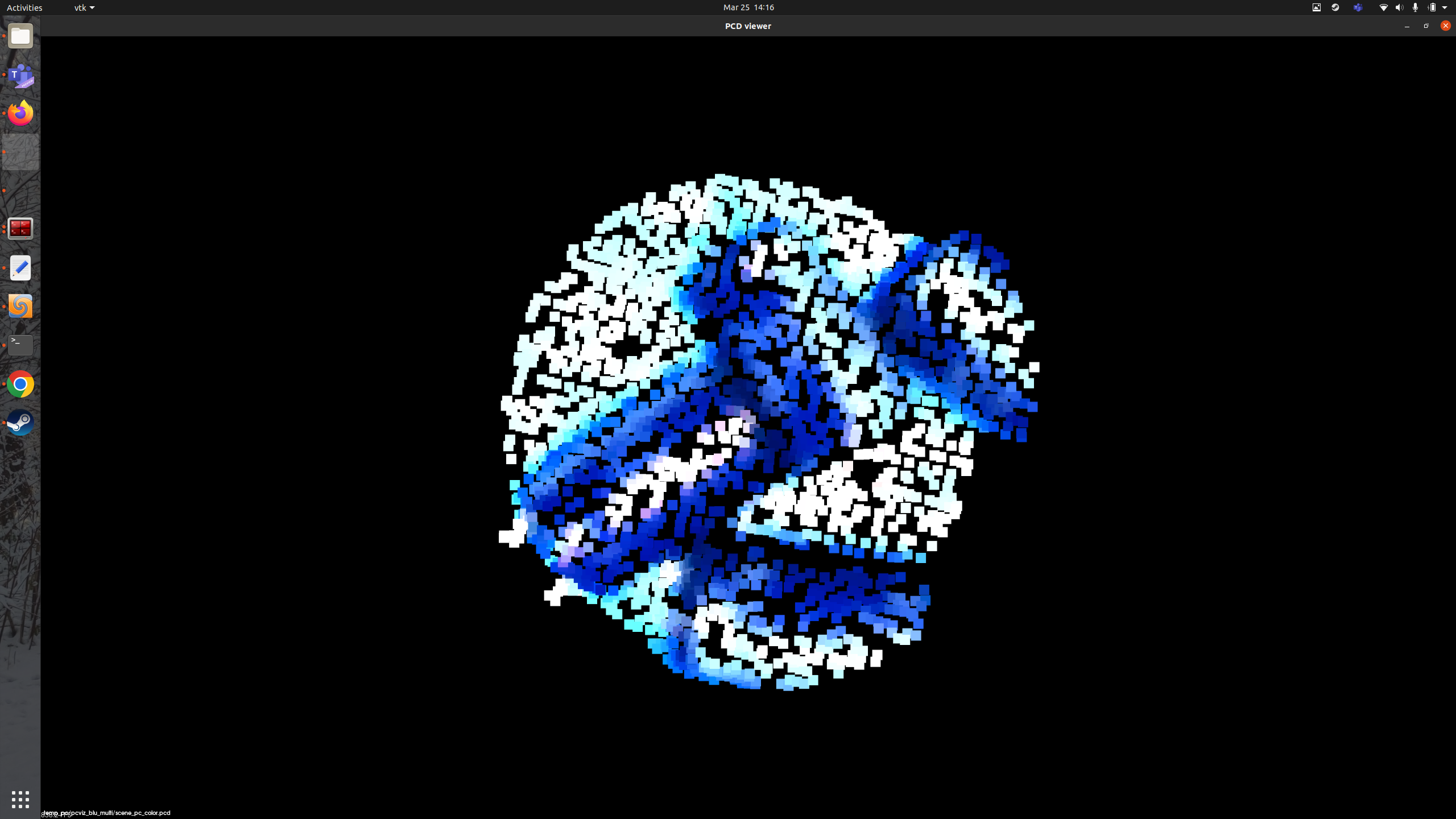} \\
            \rotatebox[origin=lB]{90}{POSE ESTIMATION} & 
            \cutpic{600}{80}{500}{160}{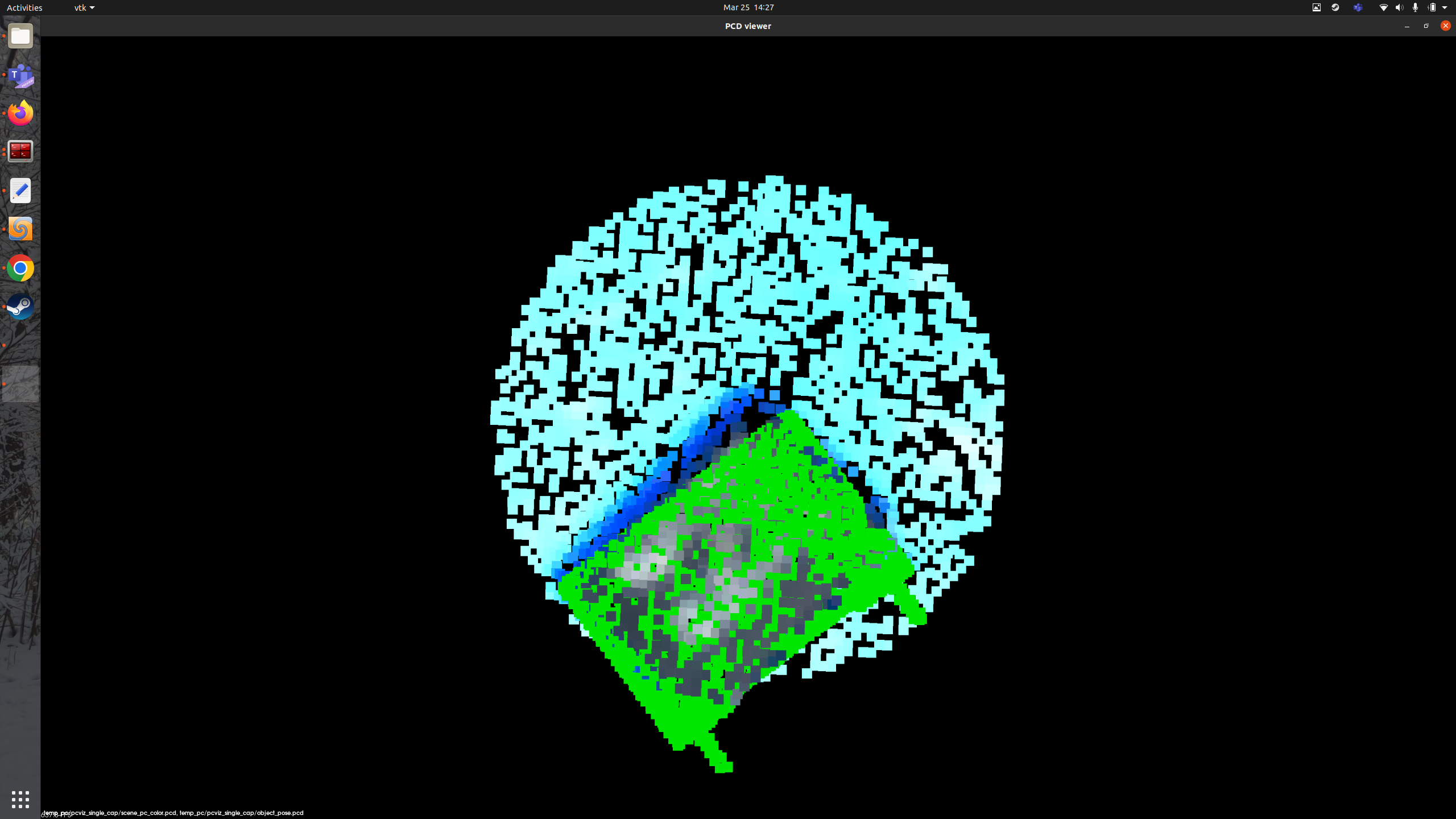} & 
            \cutpic{600}{80}{500}{160}{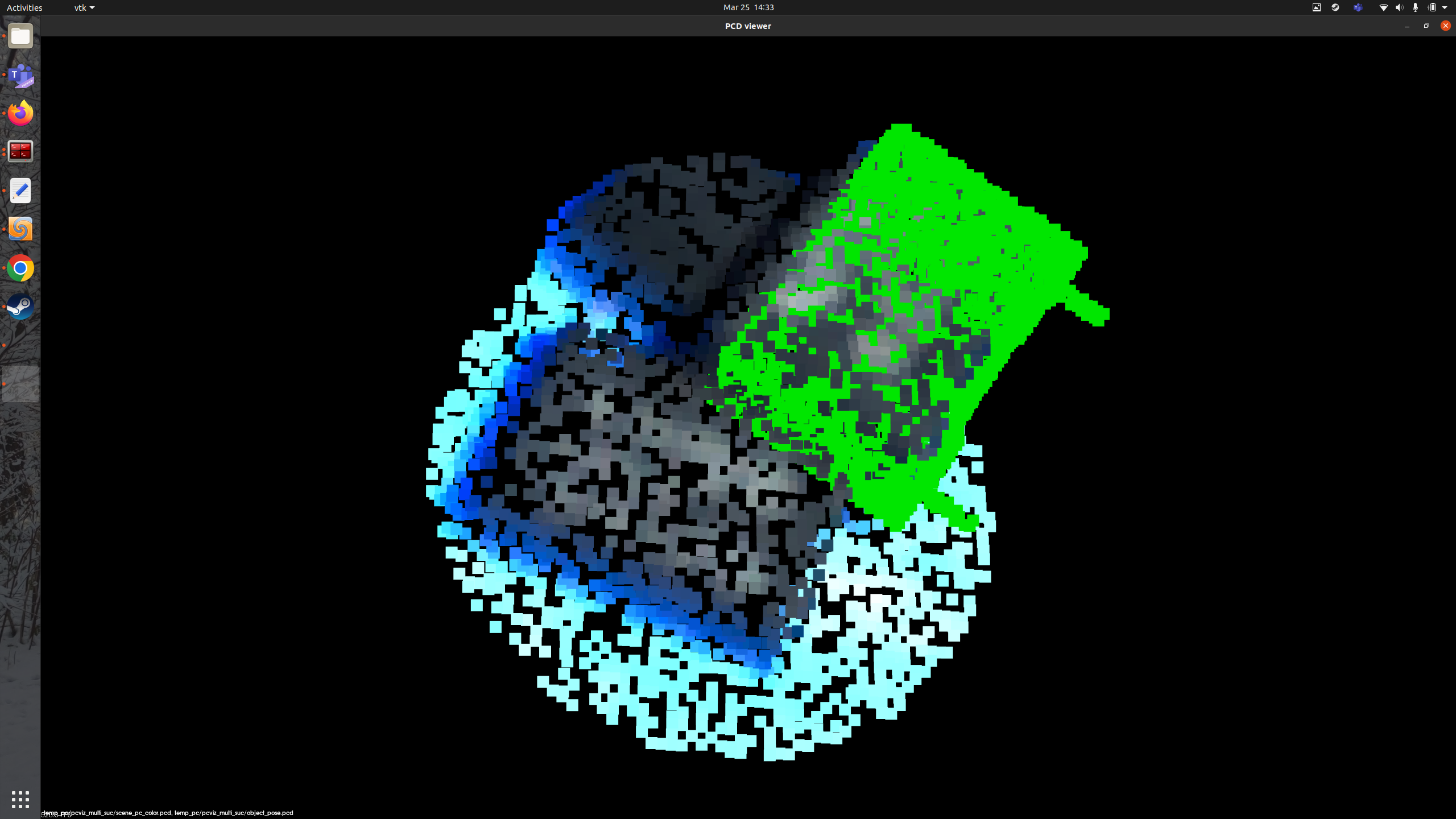} & 
            \cutpic{600}{80}{500}{160}{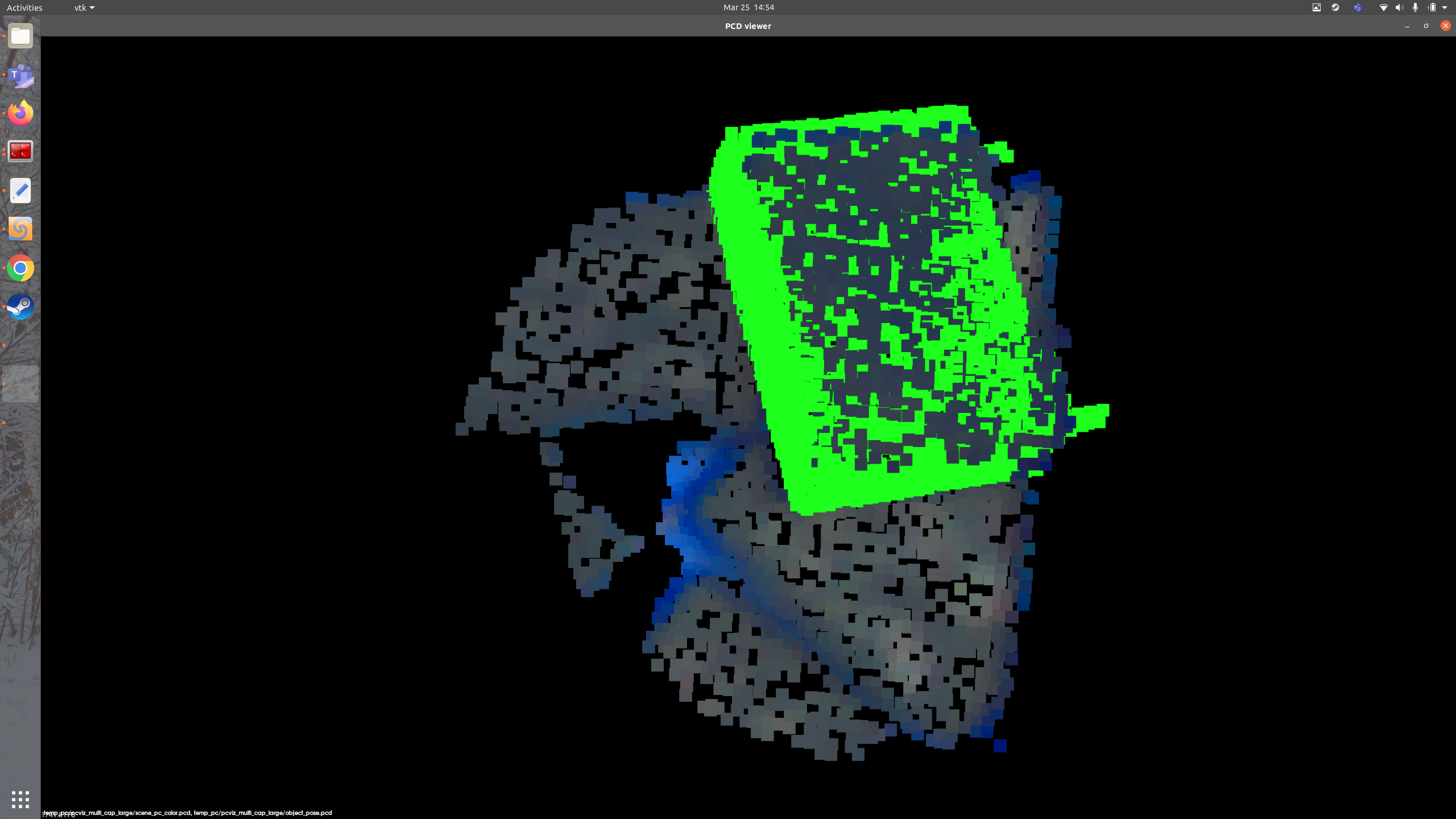} &
            \cutpic{600}{120}{500}{120}{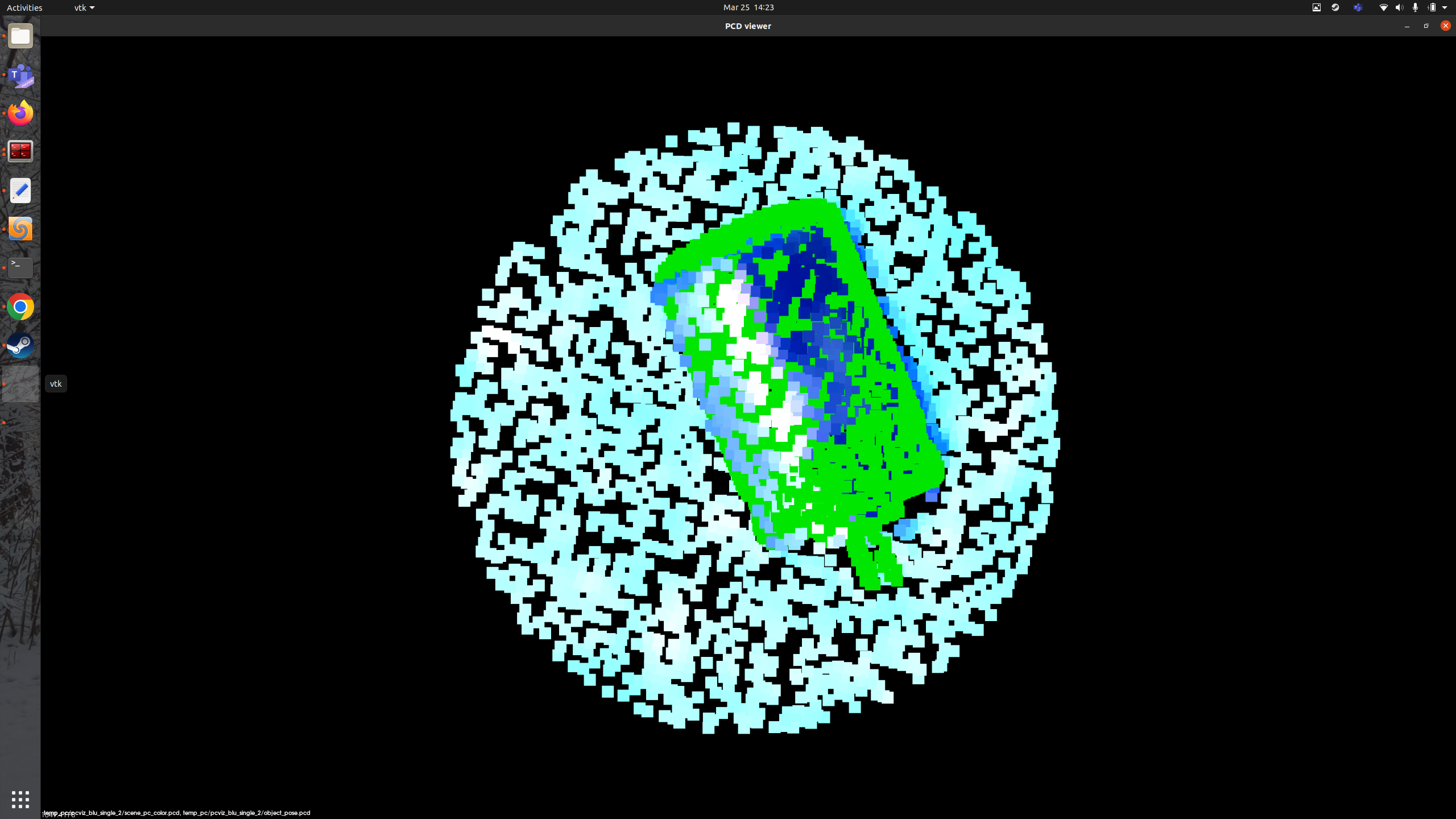} & 
            \cutpic{600}{120}{500}{120}{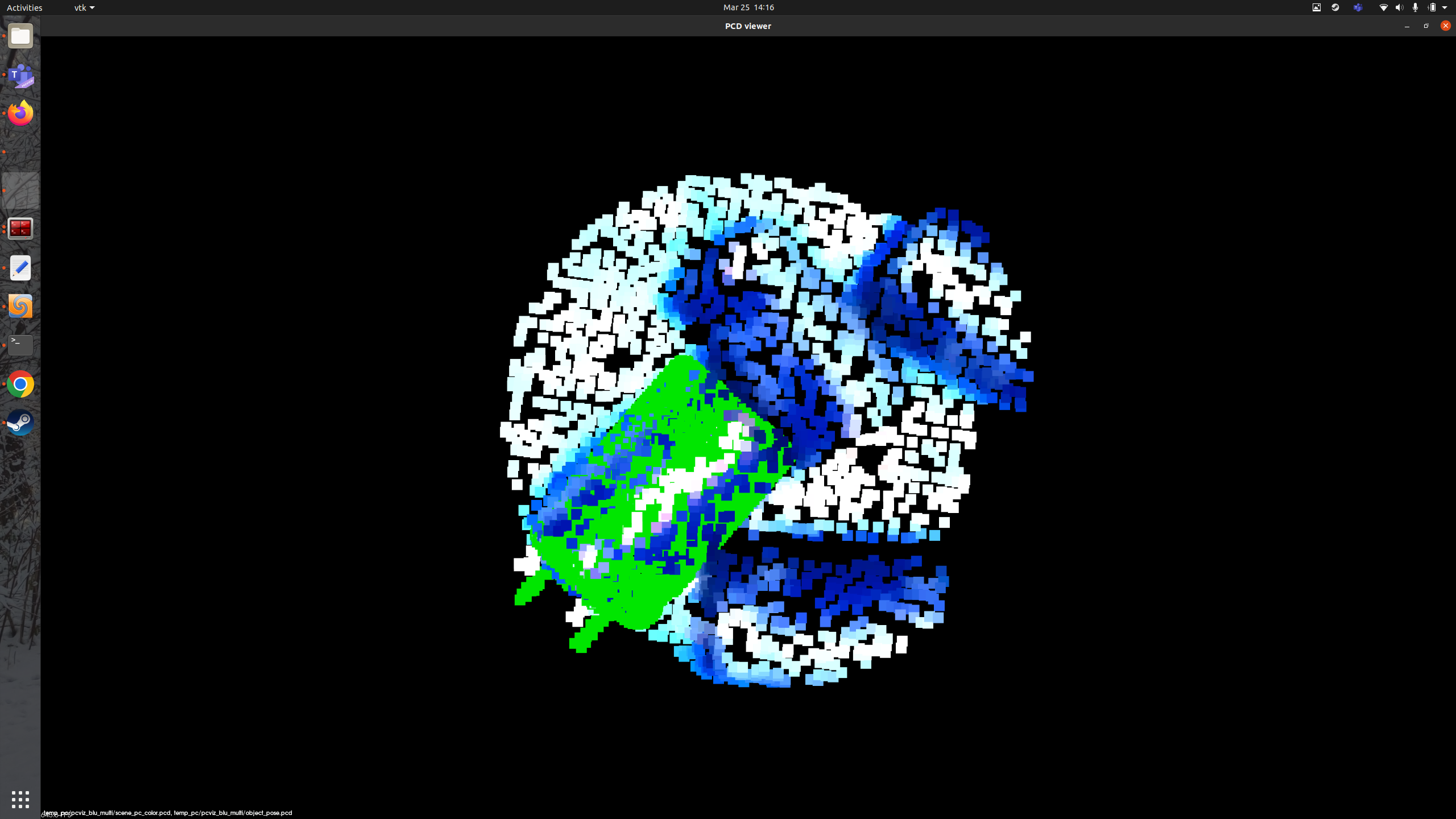} \\
       & OBJ 4 & OBJ 4 & OBJ 4 & OBJ 7 & OBJ 7 \\
    \end{tabular}
    \end{center}
    \caption{Examples of pose estimations in the real dataset. For each point cloud the scene is shown above and the pose estimation is shown underneath. }
    \label{fig:pereal}
    \vspace{-2.5mm}
\end{figure*}

\subsection{Experiments on real data}
%
%
%
To further verify the effectiveness of the developed method, experiments have been performed on real data. Real point clouds have been collected for "Obj 4" and "Obj 7" of the test set. The point clouds were collected using a Zivid2\footnote{\url{https://www.zivid.com/zivid-2}} sensor. For both objects twenty different scenes were created, with the number of objects in each scene increasing from one to twenty. For each scene, five point clouds were obtained with different viewpoints. By omitting poses not visible the final number of poses for "Obj 4" and "Obj 7" is 724 and 642, respectively. 

Our method is compared with both the classical method and a state-of-the-art pose estimation method. The state-of-the-art method is a variant of ParaPose \cite{hagelskjaer2022parapose} adapted for colorless pose estimation \cite{hagelskjaer2023off}. The method is trained according to \cite{hagelskjaer2022parapose} with synthetic data of the CAD models in the bin, similar to the training data shown in Fig.~\ref{fig:dtrainata}. 
%
The results for both objects are shown in Tab.~\ref{tab:real}. 

It is seen that our method outperforms the classic pose estimation method by a large margin, and as expected the recall is lower compared with the results on the synthetic test data.
 ParaPose \cite{hagelskjaer2022parapose} also shows very good recall, and outperforms our method by a large degree. However, when using the GPU based RANSAC our method show performance comparable with ParaPose. The increase in performance from using the GPU based RANSAC is possibly as a result of the normal vector check as shown by the decrease in recall when omitted. 

The results demonstrate that our method is able to perform pose estimation in real data, even when trained on synthetic data. We also show that the results are comparable to a state-of-the-art pose estimation method trained on the CAD model.



\begin{table}
\begin{center}
\caption{Pose estimation recall on the real data. }
\begin{tabular}{|c|c|c|c|c|c|}
\hline
 & FPFH & ParaPose & Ours & GPU & GPU w/o normal\\
\hline\hline
Obj 4      & 0.16 & 0.62 & 0.38 & 0.69 & 0.41 \\ 
Obj 7      & 0.18 & 0.59 & 0.33 & 0.77 & 0.63 \\ 
\hline
\end{tabular}
\label{tab:real}
 \vspace{-6mm}
\end{center}
\end{table}

\subsection{Run-time}
%
%
The run-time of the network was tested both with and without computing the object features at run-time. With the object features computed at run-time the processing lasts 14.9 ms. While by pre-computing the features the run-time is only 7.9 ms. The separation of object and scene feature computations is thus a significant speed-up.

Using Open3D \cite{Zhou2018} for pose estimation the full run-time is currently 84.7 ms. This is mainly related to the RANSAC which is computed on the CPU. 
By using the GPU based RANSAC the run-time is reduced to 19.9 ms. This run-time allows the processing of 50 point clouds per second.

\section{Conclusion}
This paper presents a novel method for colorless zero-shot pose estimation. 
The main contribution of the paper is the novel network structure with independent object and scene feature computation, along with a dataset of 1,500 electrical components in a homogeneous bin-picking scenario. 
The method shows very good generalizability across different objects, including out-of-class objects. The method is also demonstrated on real data where it obtains performance comparable with a state-of-the-art method trained on the test object. This proves the validity of creating object independent networks for specific scenarios, which can be useful for many real world applications.


In future work, it will be very interesting to obtain real training data to test the networks ability to learn a real scene.

The objects from the MegaPose6D \cite{labbe2022megapose} dataset could also be used to diversify the object types, and test on benchmark datasets. 

\section*{Acknowledgement}
This project was funded in part by Innovation Fund Denmark through the project MADE FAST, in part by the SDU I4.0-Lab.




\addtolength{\textheight}{-0.5cm}

\bibliographystyle{IEEEtran}
\bibliography{IEEEabrv,egbib}


\end{document}